\documentclass[10pt,twocolumn,letterpaper]{article}

\usepackage[pagenumbers]{cvpr} 

\definecolor{cvprblue}{rgb}{0.21,0.49,0.74}
\usepackage{graphicx}
\usepackage{booktabs}
\usepackage{threeparttable}
\usepackage[pagebackref,breaklinks,colorlinks,allcolors=green]{hyperref}
\usepackage{makecell}
\usepackage{algorithm}
\usepackage{algpseudocode}
\usepackage{pgffor}
\usepackage{arydshln}
\makeatletter
\def\blfootnote{\gdef\@thefnmark{}\@footnotetext}
\makeatother

\def\paperID{*****} 
\def\confName{CVPR}
\def\confYear{2026}

\title{Real-time Appearance-based Gaze Estimation for Open Domains}

\author{
Zhenhao Li\textsuperscript{1} \and Zheng Liu\textsuperscript{*1} \and Seunghyun Lee\textsuperscript{*2} \and Amin Fadaeinejad\textsuperscript{1} \and Yuanhao Yu\textsuperscript{1} \and \\
\textsuperscript{1}Huawei Technologies Canada ~ ~ 
\textsuperscript{2}University of Toronto
}

\begin{document}
\maketitle
\blfootnote{\textsuperscript{*}Equal contribution. 
Seunghyun Lee did the work during internship at Huawei.}

\begin{abstract}
Appearance-based gaze estimation (AGE) has achieved remarkable performance in constrained settings, yet we reveal a significant generalization gap where existing AGE models often fail in practical, unconstrained scenarios, particularly those involving facial wearables and poor lighting conditions. We attribute this failure to two core factors: limited image diversity and inconsistent label fidelity across different datasets, especially along the pitch axis.
To address these,
we propose a robust AGE framework that enhances generalization without requiring additional human-annotated data. First, we expand the image manifold via an ensemble of augmentation techniques, including synthesis of eyeglasses, masks, and varied lighting.
Second, to mitigate the impact of anisotropic inter-dataset label deviation, we reformulate gaze regression as a multi-task learning problem, incorporating multi-view supervised contrastive (SupCon) learning, discretized label classification, and eye-region segmentation as auxiliary objectives.
To rigorously validate our approach, we curate new benchmark datasets designed to evaluate gaze robustness under challenging conditions, a dimension largely overlooked by existing evaluation protocols.
Our MobileNet-based lightweight model achieves generalization performance competitive with the state-of-the-art (SOTA) UniGaze-H, while utilizing less than 1\% of its parameters, enabling high-fidelity, real-time gaze tracking on mobile devices.
GitHub Page: https://github.com/liszth87/GazeTorch.
\end{abstract}

\begin{figure}[tb]
     \centering
     \captionsetup[subfigure]{justification=centering, singlelinecheck=false}
     \scriptsize 
     
     \begin{subfigure}[t]{0.45\textwidth}
         \centering
         \includegraphics[width=0.85\textwidth]{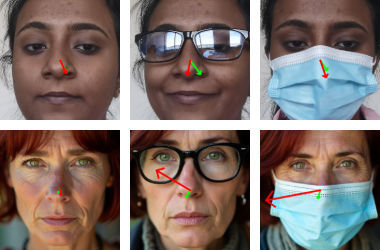}
          \caption{Red: UniGaze-H; Green: Ours} 
     \end{subfigure}%
     \hfill
     \begin{subfigure}[t]{0.48\textwidth}
         \centering
         \includegraphics[width=0.95\textwidth]{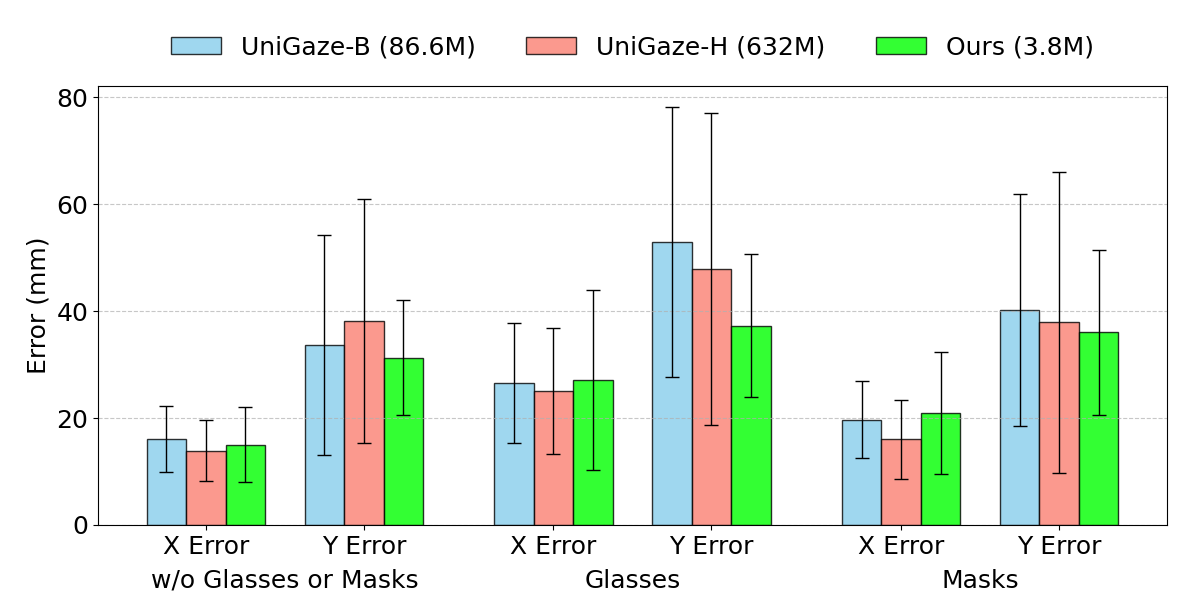}
          \caption{Evaluation on our RealGaze benchmark} 
     \end{subfigure}%
     
     \caption{
     Comparison of generalization performance between our MobileNet-based model and the SOTA UniGaze \cite{unigaze}. (a) On our benchmark datasets RealGaze (top) and ZeroGaze (bottom), UniGaze-H (red arrows) manifests significantly higher prediction variance under occlusion.
     (b) 
     Our lightweight model maintains superior robustness and manifold consistency compared to significantly larger baselines, effectively marginalizing the impact of visual alterations. Detailed analysis is provided in Section \ref{s:exp_realgaze}.
     }
     \label{fig:teaser}
\end{figure}
\section{Introduction}
\label{s:introduction}

Gaze estimation is a fundamental computer vision task:
given one or multiple images of a subject, determine their 3D gaze direction. Appearance-based gaze estimation (AGE) has garnered significant interest
because it can be deployed on ubiquitous RGB cameras found in PCs and mobile devices. However, despite rapid progress from deep learning, AGE still lags behind specialized near-infrared systems \cite{infrared} in practical deployments. A major obstacle is poor generalization (see \cref{fig:teaser}): while current models excel on standard benchmarks, they often suffer catastrophic performance degradation in unconstrained, real-world scenarios, particularly when users wear glasses or facial masks, or when illumination is challenging (\cref{s:exp_realgaze}).
Notably, we observe that this degradation is \textit{anisotropic}:
vertical (pitch) errors are consistently larger than horizontal (yaw) errors. 

We attribute this generalization gap to two interacting causes.
First, limited image diversity in existing datasets: many datasets are collected in controlled settings with few subjects and limited appearance variation, because 3D gaze labels are typically derived from 2D points‑of‑gaze via a geometric pipeline \cite{mpiigaze} that favors controlled capture to reduce labeling error.
Second, inter‑dataset label misalignment: the geometric labeling pipeline introduces systematic, dataset‑dependent biases during intermediate steps, such as camera calibration and head pose estimation.
We reveal that these errors are not uniformly distributed;
the pitch axis exhibits noticeably greater label deviation than yaw.
Consequently, the common practice of joint-dataset training \cite{unigaze, gazegla} can be actually counterproductive;
we demonstrate that na\"{i}vely pooling datasets for training can worsen pitch generalization rather then help it (\cref{s:ablation}).

A related but underexplored issue is the efficiency-accuracy tradeoff.
SOTA AGE models often rely on large backbones and heavy computation, which conflicts with the mobile, low‑latency targets where AGE is most useful.

To address these challenges, we propose a compact, generalization‑focused AGE framework free of additional human annotations. Our contributions are:
\begin{itemize}
    \item We introduce an automated augmentation pipeline that generates glasses and facial masks and simulates diverse lighting conditions to expand the image manifold and force occlusion- and illumination-robust feature learning.
    \item We reformulate AGE as a multi‑task problem that supplements regression with multi‑view SupCon learning, discretized label classification, and eye‑region segmentation. These auxiliary tasks reduce reliance on noisy regression labels, particularly along pitch, and encourage a more robust feature manifold.
    \item We curate two new benchmark datasets: RealGaze, covering diverse challenging scenarios in real-world AGE application, and ZeroGaze, a synthesized high-identity-count dataset for isolating the impact of wearables. They expose failure modes that standard cross‑domain evaluations often miss.
    \item A MobileNet‑based architecture trained with the above components matches or exceeds the generalization of heavy baselines such as UniGaze‑H while using under 1\% of their parameters, enabling high‑fidelity, low‑latency gaze tracking on mobile devices.
\end{itemize}

We validate our approach with extensive cross‑dataset experiments, per‑condition robustness analyses (glasses, masks, and poor lighting), and ablations that quantify the contribution of each component.
Results show that our augmentation pipeline and multi‑task supervision substantially improve generalization to challenging conditions, and that careful evaluation protocols are necessary to reveal true open‑domain performance.

\section{Related Work}
\label{s:related}

\subsection{Appearance-based Gaze Estimation (AGE)}

Gaze estimation has evolved from early geometric eye modeling \cite{eyeball, eyetab} and manual feature analysis \cite{local, tip} to end-to-end deep networks. Currently, AGE models following the face-to-gaze paradigm established by MPIIFaceGaze \cite{mpiiface, normalization} represent the de-facto standard. Despite architectural advances, AGE models still face a substantial generalization bottleneck, largely due to the scarcity of reliable labeled training data \cite{unigaze}. This has spurred a shift from single-domain optimization toward cross-domain generalization, primarily via test-time adaptation and representation learning. Test‑time adaptation fine‑tunes models on target data, either with labels \cite{faze, differential, gaze322} or without \cite{crga, ruda, unrega, metaprompt}. However, obtaining accurate labels at test time is difficult, and on‑device optimization is often impractical for mobile deployment. Representation learning seeks domain‑invariant features using semi‑ or self‑supervised objectives \cite{omnigaze, puregaze, unigaze}, but these settings remain tethered to the quality of the labeled data used for the feature-to-gaze mapping. To improve training signals, some data‑centric approaches correct inter‑dataset label discrepancies \cite{gazegla} or synthesize high‑fidelity labeled images via computer graphics pipelines \cite{gazegene}. Recent work also explores weak supervision from vision-language models \cite{dcgaze, clipgaze}. Despite these efforts, few works explicitly address the anisotropic nature of inter‑dataset label deviation, or the performance degradation induced by occlusions and poor illumination in real-world deployment.

\subsection{Supervised Contrastive (SupCon) Learning}

Our work utilizes SupCon \cite{supcon} to manipulate the feature space by fully harnessing both gaze labels and auxiliary attributes beyond gaze (e.g., source datasets, glasses and mask states).
Unlike standard self-supervised contrastive learning, SupCon leverages label information to group similar samples together in the embedding space. Formally, given a multi-viewed batch $B=\{\mathbf{z}_1, ..., \mathbf{z}_N\}$ of normalized feature vectors, define $P(i) = \{p \in \{1, ..., N\} \backslash \{i\} : M_{ip} = 1 \}$ as the index set of positives for sample $i$, where $M_{ij}=1$ indicates a positive pair.
The SupCon loss for the batch $B$ is 
{\small
\begin{equation}
\label{eq:supcon}
    L^S (B) = \sum_{i=1}^N \frac{-1}{|P(i)|} \sum_{p \in P(i)} \log \frac{\exp(\mathbf{z}_i \cdot \mathbf{z}_p / \tau_S)}{ \sum_{q \neq i} \exp(\mathbf{z}_i \cdot \mathbf{z}_q / \tau_S)}, 
\end{equation}}
pulling features from positive pairs closer together than negative pairs.
$\tau_S$ is a preset temperature parameter that controls the alignment strength.

\begin{table*}[tb]
  \caption{ 
  Summary of representative AGE datasets.
  The datasets selected for our training pipeline are \textbf{bolded}.
  We define the symbols $D_*$ for simplicity. 
  }
  \label{tab:dataset_size}
  \centering
  \resizebox{0.9\textwidth}{!}{%
  \begin{threeparttable}
  \begin{tabular}{@{}c | ccc | cc@{}}
    \toprule
    \textbf{Dataset} & \multicolumn{3}{c}{\textbf{Diversity}} & 
    Image & Label \\
     & Subject\# & Sample\# & Gaze Range$^\dagger$ (pitch $\times$ yaw)~  & Quality & Fidelity$^*$ \\
    \midrule
    $D_E$: EYEDIAP (FT sessions excluded) \cite{eyediap} & 14 \cite{crga}$^\ddagger$ & 15K &$[5^\circ, 21^\circ] \times [-17^\circ, 16^\circ ]$ & Low & Medium \\
    $D_C$: \textbf{GazeCapture} \cite{gazecapture} & 1474 & 2M & $[-24^\circ, 9^\circ] \times [-21^\circ, 21^\circ]$ & Medium & Medium \\ 
    $D_M$: MPIIFaceGaze \cite{mpiiface} & 15 & 45K & $[-24^\circ, 0^\circ] \times [-20^\circ, 20^\circ]$ & Medium & Medium \\
    $D_{360}$: Gaze360 \cite{gaze360} & 238 & 100K \cite{crga} & $[-62^\circ, 15^\circ] \times [-75^\circ, 72^\circ]$ & Low & Low \\ 
    $D_X$: \textbf{ETH-XGaze} \cite{xgaze} & 80 & 760K & $[-65^\circ, 56^\circ] \times [-86^\circ, 82^\circ]$ & High & High  \\ 
    $D_N$: \textbf{GazeGene} \cite{gazegene} & 56 & 1M & $[-59^\circ, 53^\circ] \times [-78^\circ, 78^\circ]$ & Medium & Medium \\ 
  \bottomrule
  \end{tabular}
  \begin{tablenotes}
    \footnotesize  
    \item[$\dagger$] We exclude the 5\% of samples with the most extreme gaze labels as outliers.
    The pitch axis and yaw axes point top and left, respectively.
    $^\ddagger$ We follow CRGA \cite{crga} to pre-process the datasets. $^*$ See Appendix for details.
    \end{tablenotes}
  \end{threeparttable}
  }
\end{table*}

\section{Data Constraints in AGE}
\label{s:labels}

The performance envelope of AGE models is fundamentally dictated by the diversity and fidelity of available training data. Unlike related domains such as face recognition, current gaze datasets remain small in sample and subject count along with environmental variation (\cref{tab:dataset_size}), largely because obtaining precise 3D gaze labels is inherently difficult. This section outlines two systemic constraints of existing datasets that motivate our framework.

\subsection{The Diversity-Fidelity Tradeoff}
Standard gaze labeling requires subjects to fixate on known stimuli under controlled conditions. This setup ensures geometric accuracy but severely restricts identity diversity, environmental variation, and the presence of real‑world artifacts such as wearables or challenging illumination. Crowdsourced datasets like GazeCapture $D_C$ \cite{gazecapture} partially address scale, yet the relaxed capture protocol introduces substantial noise (\eg, motion blur, closed eyes, and distracted subjects) and the small‑device setup yields a narrower point-of-gaze (PoG) distribution than laboratory datasets. Synthetic datasets such as GazeGene $D_N$ \cite{gazegene} expand identity diversity, but they struggle to reproduce fine‑grained real‑world effects, including complex ocular reflections, sensor‑specific imaging artifacts, and diverse occlusion patterns. As a result, no existing dataset simultaneously offers high fidelity, broad appearance variation, and realistic occlusion/illumination conditions.

\subsection{Anisotropic Inter-dataset Label Deviation}
While inter-dataset label deviation is widely acknowledged in the literature \cite{crga, gazegla},
their directional structure is rarely quantified. We conduct a cross‑dataset perceptual study to measure label consistency across the four largest labeled datasets $\{ D_X,D_N,D_C,D_{360}\}$. Annotators compare 400 image pairs matched by gaze and head pose labels (within $4^\circ$) and judge whether the two samples depict the same gaze direction, evaluated separately for pitch and yaw.

Results in \cref{tab:user_study} reveal a pronounced anisotropy: pitch labels exhibit substantially lower perceptual consistency than yaw across all dataset combinations.
This suggests that pitch labels are inherently more susceptible to systematic errors during the PoG-to-3D mapping process (see Appendix for further discussion). Consistency also varies by source; for example, Gaze360 $D_{360}$ \cite{gaze360} contains many low‑resolution or heavily degraded samples where gaze direction is difficult to discern, leading to markedly lower consistency than the remaining datasets.
Based on these findings, we prioritize $\{ D_X,D_N,D_C\}$ for joint training,
while designing our multi‑task objectives to explicitly mitigate unreliable pitch supervision, particularly from $D_N$ and $D_C$, whose perceptual pitch consistency is noticeably lower than that of $D_X$.

\begin{table}[tb]
    \caption{
    Quantitative and qualitative analysis of inter-dataset label consistency. Perceived consistency rates on pitch and yaw axes reveal a significant discrepancy, confirming the presence of anisotropic inter-dataset label deviation.
    }
    \label{tab:user_study}
      \centering
      \scriptsize
      \renewcommand{\arraystretch}{1.0} 
      \setlength{\tabcolsep}{4pt} 
      \begin{subtable}[t]{0.48\columnwidth}
        \centering
        \caption{Pitch}
        \label{tab:pitch}
        \resizebox{\linewidth}{!}{%
        \begin{tabular}{@{}r | cccc@{}}
        \toprule
                   & $D_X$ & $D_N$ & $D_C$ & $D_{360}$ \\ \midrule
        $D_X$      & ---   & 0.750 & 0.646 & 0.440 \\
        $D_N$      & 0.750 & ---   & 0.529 & 0.463 \\
        $D_C$      & 0.646 & 0.529 & ---   & 0.323 \\
        $D_{360}$  & 0.440 & 0.463 & 0.323 & ---   \\ \midrule
        Average    & 0.612 & 0.581 & 0.499 & 0.409 \\ \bottomrule
        \end{tabular}}
      \end{subtable}
      \hfill
      \begin{subtable}[t]{0.48\columnwidth}
        \centering
        \caption{Yaw}
        \label{tab:yaw}
        \resizebox{\linewidth}{!}{%
        \begin{tabular}{@{}r | cccc@{}}
        \toprule
                   & $D_X$ & $D_N$ & $D_C$ & $D_{360}$ \\ \midrule
        $D_X$      & ---   & 0.859 & 0.768 & 0.573 \\
        $D_N$      & 0.859 & ---   & 0.729 & 0.653 \\
        $D_C$      & 0.768 & 0.729 & ---   & 0.532 \\
        $D_{360}$  & 0.573 & 0.593 & 0.532 & ---   \\ \midrule
        Average    & 0.733 & 0.747 & 0.676 & 0.586 \\ \bottomrule
        \end{tabular}}
      \end{subtable}
\end{table}
\begin{figure}[t]
      \hfill
        \centering
        \captionsetup[subfigure]{justification=centering, singlelinecheck=false}
        \begin{subfigure}{0.23\columnwidth}
            \centering
            \includegraphics[width=1.8cm]{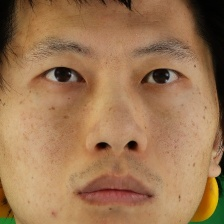}
            \caption*{$D_X$}
        \end{subfigure}
        \hfill
        \begin{subfigure}{0.23\columnwidth}
            \centering
            \includegraphics[width=1.8cm]{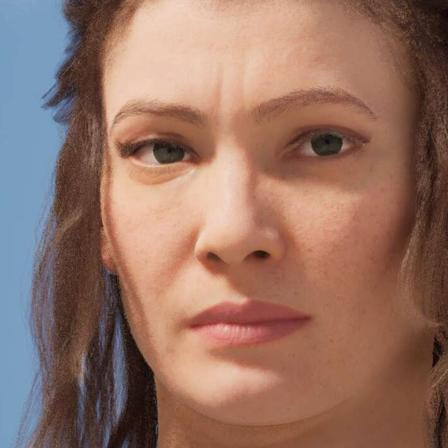}
            \caption*{$D_N$}
        \end{subfigure}
        \hfill 
        \begin{subfigure}{0.23\columnwidth}
            \centering
            \includegraphics[width=1.8cm]{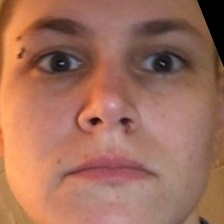}
            \caption*{$D_C$}
        \end{subfigure}
        \hfill
        \begin{subfigure}{0.23\columnwidth}
            \centering
            \includegraphics[width=1.8cm]{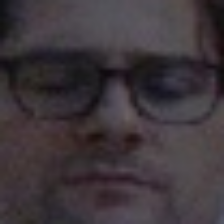}
            \caption*{$D_{360}$}
        \end{subfigure}
        \caption{Qualitative comparison of samples from different datasets with near-identical annotations; visually divergent gaze directions, particularly in the pitch dimension, illustrate the inherent unreliability of cross-dataset vertical ground truths.}
        \label{fig:user_study}
\end{figure}

\begin{figure*}[tb]
  \centering
  \includegraphics[width=0.9\linewidth]{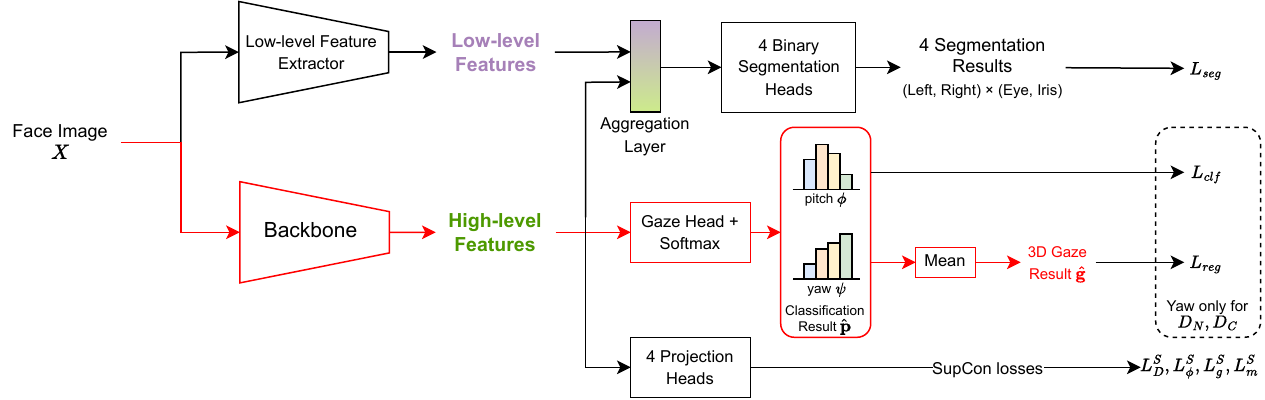}
  \caption{
  Framework of the proposed multi-task learning architecture. 
  Red blocks indicate the streamlined architecture during inference. 
  }
  \label{fig:archi}
\end{figure*}

\section{Methodology}
\label{s:method}
Given labeled data $D = \{X_i, \mathbf{g}_i\}$ consisting of
face images $X$ and corresponding 3D gaze labels in Euler angles $\mathbf{g} = (\phi, \psi)$,
our objective is to learn a mapping that generalizes reliably across domains. 
Here, $\phi$ and $\psi$ represent the pitch and yaw angles, respectively, and $D= D_X \cup D_N \cup D_C$.
To overcome the data constraints identified earlier, we expand the limited image diversity of existing datasets through an automated augmentation pipeline (\cref{s:aug})
and then introduce a multi-task learning paradigm (\cref{s:joint}) that incorporates auxiliary objectives, including discretized label classification, eye and iris segmentation, and multi-view SupCon learning,
to regularize the feature space.
We further apply label resampling to counter non‑uniform gaze label distributions, and selectively drop pitch‑related terms for $D_N$ and $D_C$, to mitigate the impact of anisotropic inter-dataset label deviation.
The detailed model architecture is subsequently presented in \cref{s:archi}.

\subsection{Automated Data Augmentation Pipeline}
\label{s:aug}

\begin{figure}[t] 
     \centering
     \captionsetup[subfigure]{justification=centering, singlelinecheck=false, labelsep=newline}
     \scriptsize 
     
     \begin{subfigure}[t]{0.09\textwidth}
         \centering
         \includegraphics[width=\textwidth]{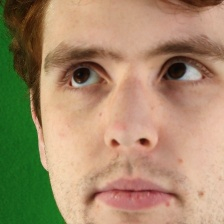}
         \caption{Original}
     \end{subfigure}%
     \hfill
     \begin{subfigure}[t]{0.09\textwidth}
         \centering
         \includegraphics[width=\textwidth]{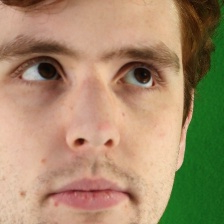}
         \caption{Flip}
     \end{subfigure}%
     \hfill
     \begin{subfigure}[t]{0.09\textwidth}
         \centering
         \includegraphics[width=\textwidth]{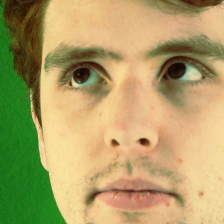}
         \caption{Color Jitter}
     \end{subfigure}%
     \hfill
     \begin{subfigure}[t]{0.09\textwidth}
         \centering
         \includegraphics[width=\textwidth]{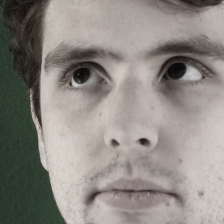}
         \caption{Desaturation}
     \end{subfigure}%
     \hfill
     \begin{subfigure}[t]{0.09\textwidth}
         \centering
         \includegraphics[width=\textwidth]{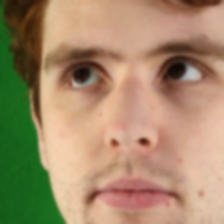}
         \caption{Blur}
     \end{subfigure}

     \vspace{1em} 

     \begin{subfigure}[t]{0.09\textwidth}
         \centering
         \includegraphics[width=\textwidth]{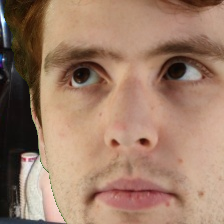}
         \caption{\begin{tabular}[t]{@{}c@{}}\scalebox{1}{Background} \\ \scalebox{1}{Replacement}\end{tabular}}
     \end{subfigure}%
     \hfill
     \begin{subfigure}[t]{0.09\textwidth}
         \centering
         \includegraphics[width=\textwidth]{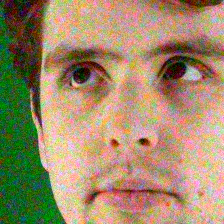}
         \caption{\begin{tabular}[t]{@{}c@{}}\scalebox{1}{Sensor} \\ \scalebox{1}{Noise}\end{tabular}}
     \end{subfigure}%
     \hfill
     \begin{subfigure}[t]{0.09\textwidth}
         \centering
         \includegraphics[width=\textwidth]{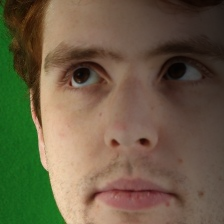}
         \caption{\begin{tabular}[t]{@{}c@{}}\scalebox{1}{Illumination} \\ \scalebox{1}{Perturbation}\end{tabular}}
     \end{subfigure}%
     \hfill
     \begin{subfigure}[t]{0.09\textwidth}
         \centering
         \includegraphics[width=\textwidth]{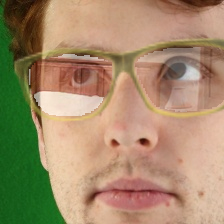}
         \caption{\begin{tabular}[t]{@{}c@{}}\scalebox{1}{Glasses} \\ \vphantom{X}\end{tabular}}
     \end{subfigure}%
     \hfill
     \begin{subfigure}[t]{0.09\textwidth}
         \centering
         \includegraphics[width=\textwidth]{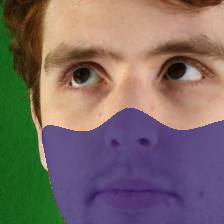}
         \caption{\begin{tabular}[t]{@{}c@{}}\scalebox{1}{Masks} \\ \vphantom{X}\end{tabular}}
     \end{subfigure}

     \caption{Overview of the automated data augmentation pipeline. During training, we stochastically combine these methods for each sample to expand the training manifold.}
     \label{fig:aug}
\end{figure}

The scarcity of high-fidelity labels limits the coverage of existing AGE datasets, often leading to model failure in unconstrained scenarios involving complex occlusions and lighting.
We therefore construct a suite of augmentations that expand the training manifold toward realistic open‑domain conditions (\cref{fig:aug}). These augmentations combine standard transformations with domain‑specific synthesis tailored to AGE, as specified below.

\paragraph{Background Diversification}
Laboratory and synthetic datasets such as $D_X$ and $D_N$ often contain static, dataset‑specific backgrounds. To prevent spurious correlations, we apply portrait matting \cite{mediapipe} to extract the subject and replace the background with random indoor scenes from the MIT Indoor dataset \cite{MITIndoor}, increasing environmental variability (\cref{fig:aug}(f)).

\paragraph{Realistic Sensor Noise}
While synthetic data from $D_N$ provides high resolution, it lacks the stochastic noise characteristic of real CMOS sensors. Rather than applying standard global Gaussian noise, we introduce a heuristic noise model that  approximates sensor-level artifacts (\cref{fig:aug}(g)).
This reduces the domain gap between synthetic and real-world images and is equally applicable to real datasets.

\paragraph{Illumination Perturbation}
As shown in \cref{fig:aug}(h), we simulate directional light sources by overlaying linear gradients with randomized opacity, color tone, and spatial orientation.
This forces the model to focus on geometric ocular structures rather than pixel intensities, which may be skewed by illumination.

\begin{figure*}[t]
     \centering
     \captionsetup[subfigure]{justification=centering, singlelinecheck=false}
     \scriptsize 
     
     \begin{subfigure}[t]{0.3\textwidth}
         \centering
         \includegraphics[height=5cm]{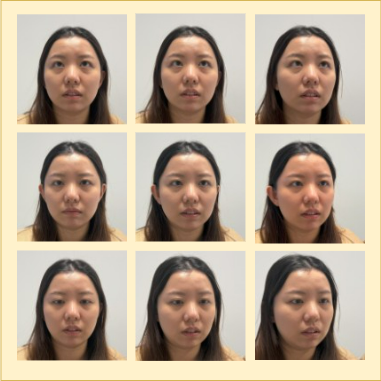}
         \caption{}
     \end{subfigure}%
     \hfill
     \begin{subfigure}[t]{0.3\textwidth}
         \centering
         \includegraphics[height=5cm]{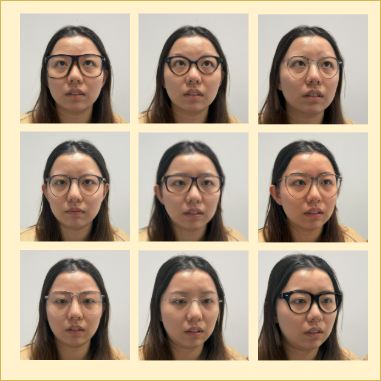}
         \caption{}
     \end{subfigure}%
     \hfill
     \begin{subfigure}[t]{0.4\textwidth}
         \centering
         \includegraphics[width=5.8cm]{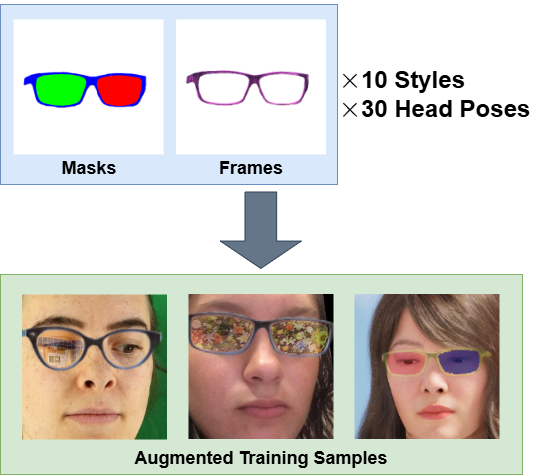}
         \caption{}
     \end{subfigure}

     \caption{
     Pipeline for pose-consistent eyeglasses template generation. (a) Original face images, with 30 discrete head poses.
     (b) GlassesGAN outputs, featuring diverse frame styles.
     (c) Extracted glasses templates and examples of augmented training samples.}
     \label{fig:glasses_aug}
\end{figure*}

\paragraph{Pose-Consistent Eyeglasses Synthesis}
Synthesizing glasses requires geometric alignment with the subject's head pose to remain visually plausible.
To achieve this efficiently, we utilize GlassesGAN \cite{glassesgan} to generate an offline library of 300 glasses templates, covering 10 frame styles across a grid of 30 discrete head poses (ranging from $[-30^\circ, 30^\circ]$ in pitch and $[0^\circ, 27^\circ]$ in yaw), as shown in \cref{fig:glasses_aug}.
During training, 
we retrieve the template whose head pose is closest to that of the sample (mirroring templates for negative yaw), align it using facial landmarks, and randomize frame size, color, and opacity.
To mimic real lens reflections, we alpha‑blend indoor scene textures onto the lens regions, forcing the model to ``see through'' reflection artifacts and rely on stable ocular cues.

\paragraph{Synthesizing Mask Occlusion}
We simulate masks by filling the lower‑face region, defined by landmarks on the nose, cheeks, and jawline, with random solid colors or textures (\cref{fig:aug}(j)). This preserves overall facial geometry while reducing the model's reliance on fine‑grained lower‑face appearance, encouraging it to focus on periocular cues that are more reliable for gaze inference.

\subsection{AGE as a Multi-task Learning Problem}
\label{s:joint}

A na\"{i}ve approach to AGE is the direct minimization of regression loss (\eg, $\ell_1$ or angular loss). However, we observe that this often leads to ``mean-collapse,'' where the model's predictions gravitate toward the dataset's mean gaze vector.
This failure arises from two factors:
non-uniform label distribution and inter-dataset label deviation.
To learn a robust gaze feature space that resists these biases,
we reformulate AGE as a multi-task learning problem, incorporating the following auxiliary objectives.

\subsubsection{Resampling and Classification with Discretized Labels}
\label{s:clf}
To eliminate data bias, we define a gaze range of interest
$I: [\phi_{min}, \phi_{max}] \times [\psi_{min}, \psi_{max}]$ covered by all three datasets.
We partition $I$ into an $n_\phi \times n_\psi$ grid, with sizes $s_\phi = \frac{\phi_{max}-\phi_{min}}{n_\phi}$ and $s_\psi = \frac{\psi_{max}-\psi_{min}}{n_\psi}$.
During each training epoch, we perform a stratified resampling by drawing a consistent number of samples per bin and per dataset. This dual balancing strategy ensures an approximately uniform label distribution across $I$ while simultaneously equalizing the sample frequency between $D_X$, $D_N$ and $D_C$. 

Each gaze label $(\phi, \psi)$ is assigned a discrete label $(c_\phi = \lfloor \frac{\phi - \phi_{min}}{s_\phi} \rfloor, c_\psi = \lfloor \frac{\psi - \psi_{min}}{s_\psi} \rfloor)$,
enabling an auxiliary classification task. 
For pitch, the model outputs an $n_\phi$-dimensional probability vector $\hat{\mathbf{p}}_\phi = (\hat p_1, ..., \hat p_{n_\phi})$ via softmax,
and calculates the final estimate as the expectation over bin centroids
$\hat \phi = \Sigma_{i=1}^{n_{\phi}} \hat p_i \phi_i$,
where $\phi_i = \phi_{min} + (i - \frac{1}{2}) s_\phi$ is the centroid of bin $i$.
We supervise $\hat{\mathbf{p}}_\phi$ and $\hat \phi$ via a joint cross-entropy loss $L_{clf}$ and an $\ell_1$ regression loss $L_{reg}$.
This formulation prevents mean-collapse by forcing the model to distinguish between discrete gaze zones. We further sharpen $\hat{\mathbf{p}}_\phi$ using a low temperature $\tau = 0.5$ in the softmax, penalizing dispersed predictions without requiring an explicit variance loss \cite{mean-var}. The same formulation applies to the yaw dimension.

\subsubsection{Attenuating Supervision from Unreliable Labels}
\label{s:weak}

As established in \cref{s:labels}, inter-dataset label deviation is highly anisotropic, with pitch labels exhibiting significantly higher variance than yaw.
Our empirical analysis (\cref{s:ablation}) confirms that this deviation is functionally destructive: training on a na\"{i}ve union of all pitch labels yields worse generalization than training on $D_X$ pitch labels alone.
While GLA \cite{gazegla} attempts to calibrate dataset offsets by iteratively training and comparing models across dataset combinations, it incurs prohibitive overhead. 
In contrast, we propose a more efficient alternative: we selectively discard $L_{reg}$ and $L_{clf}$ for pitch labels from $D_N$ and $D_C$, where pitch labels exhibit lower fidelity.
Instead, these ``noisy'' labels contribute through a pitch-aware SupCon loss $L_\phi^S$, where positive pairs are defined by pitch differences within $s_\phi$.
This shifts supervision from absolute coordinates to relative manifold alignment, allowing the model to learn meaningful vertical structure while remaining robust to systematic pitch offsets.

\subsubsection{Multi-view SupCon Learning}
\label{s:supcon}

The diversity introduced by our augmentation pipeline enables multi-view SupCon learning to enforce invariance to gaze-irrelevant factors. For each training image, we generate $n$ independent augmentations to form multiple views.
We introduce four specific SupCon terms:
1)
pitch contrastive term $L^{S}_{\phi}$, anchoring the pitch feature manifold, as described above;
2)
dataset-invariance term $L^{S}_{D}$, with positive pairs being samples from different sources, preventing the backbone from encoding dataset-specific signatures;
3) glasses-invariance term $L^S_g$, with positive pairs being samples
with different glasses states, encouraging robustness to glasses occlusion and reflections; and 
4) mask-invariance term $L^S_m$, similarly enforcing invariance to lower-face occlusion.
We adopt $n=4$ to balance computational efficiency with the need for a sufficiently dense set of positive pairs associated with $L^S_\phi$.
To further enrich view diversity, we apply horizontal flipping (\cref{fig:aug}(b)) only to even‑indexed views, doubling yaw variation by negating yaw labels for mirrored samples.

Intuitively, one can also impose regularization on the $n$ views of a sample to enforce output-level consistency, exemplified by the symmetry loss \cite{gaze360, metaprompt} which constrains mirrored image pairs to exhibit opposite yaw angles.
But we find that directly constraining the regression head often re‑introduces mean‑collapse.
In contrast, feature‑level contrastive regularization shapes a more stable and discriminative gaze manifold, mitigating collapse while preserving fine‑grained directional structure.

\subsubsection{Robust Eye and Iris Segmentation}
\label{s:seg}

\begin{figure}[t] 
     \centering
     \captionsetup[subfigure]{justification=centering, singlelinecheck=false}
     \scriptsize 
     \begin{subfigure}[t]{0.23\columnwidth}
         \centering
         \includegraphics[width=\linewidth]{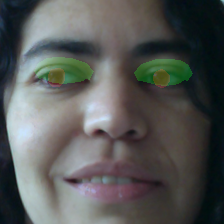}
         \caption{Ground truth} 
     \end{subfigure}%
     \hfill
     \begin{subfigure}[t]{0.23\columnwidth}
         \centering
         \includegraphics[width=\linewidth]{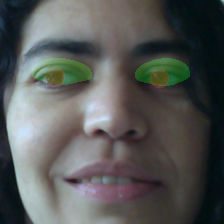}
         \caption{Result} 
     \end{subfigure}%
     \hfill
     \begin{subfigure}[t]{0.23\columnwidth}
         \centering
         \includegraphics[width=\linewidth]{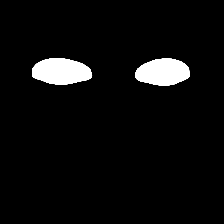}
         \caption{Result (eyes)} 
     \end{subfigure}%
     \hfill
     \begin{subfigure}[t]{0.23\columnwidth}
         \centering
         \includegraphics[width=\linewidth]{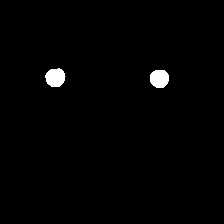}
         \caption{Result (irises)} 
     \end{subfigure}%
     \caption{
     Results of eye and iris segmentation produced by our MobileNet-based model. In (a) and (b), green regions denote the eye, while red denotes the iris.
     }
     \label{fig:seg}
\end{figure}

Discerning the fine-grained geometry of the eye and iris is essential for accurate AGE, yet these features are easily obscured by wearables or poor lighting.
We introduce four binary segmentation tasks (left/right eye and iris) as auxiliary objectives to anchor the representation of ocular appearance. 
Since segmentation masks are unavailable in most datasets,
we utilize MediaPipe landmarks \cite{mediapipe} to generate eye-region ground truths and an image-processing pipeline to estimate iris masks (detailed at Appendix). Supervision via the Dice loss ($L_{seg}$) ensures that the backbone extracts stable, high-fidelity ocular cues even in challenging in-the-wild scenarios (\cref{fig:seg}).

\subsection{Model Architecture}
\label{s:archi}

The overall training objective combines regression, discretized classification, segmentation, and four SupCon terms:
\begin{equation} \label{eq:loss}
\begin{split}
    L &= L_{reg} + \lambda_{clf} L_{clf} + \lambda_{seg} L_{seg} \\
        &+ \lambda_D L^{S}_{D} + \lambda_\phi L^{S}_{\phi} + \lambda_g L^{S}_{g} + \lambda_m L^{S}_{m}.
\end{split}
\end{equation}
To support these objectives, we augment the backbone with a lightweight segmentation branch and separate projection heads for each SupCon term (\cref{fig:archi}).
Following BiSeNet-v2 \cite{bisenet}, we fuse high-level semantic features from the backbone with low-level features from a shallow CNN via a BiSeNet aggregation layer. This aggregated feature serves as the input for the segmentation heads.

Although the framework is compatible with arbitrary backbones, our primary lightweight model adopts MobileNet‑v2 \cite{mobilenet} enhanced with Coordinate Attention (CA) \cite{ca}. 
This allows for an expanded receptive field with minimal computational cost, enabling real-time inference on commodity mobile devices.

\begin{figure}[t] 
     \centering
     \captionsetup[subfigure]{justification=centering, singlelinecheck=false}
     \scriptsize 
     
     \begin{subfigure}[t]{0.25\columnwidth}
         \centering
         \includegraphics[width=1.8cm]{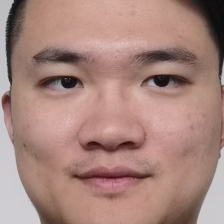}
         \caption*{$a$: $\emptyset$}
     \end{subfigure}
     \hspace{1mm}
     \begin{subfigure}[t]{0.25\columnwidth}
         \centering
         \includegraphics[width=1.8cm]{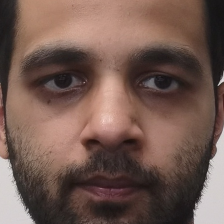}
         \caption*{$b$: $\{O\}$}
    \end{subfigure}%
     \hspace{1mm}
     \begin{subfigure}[t]{0.25\columnwidth}
         \centering
         \includegraphics[width=1.8cm]{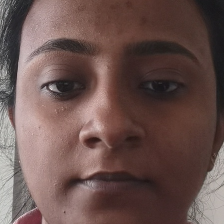}
         \caption*{$c$: $\{ S \}$}     
     \end{subfigure}%
    \vspace{0.1mm} 
     \begin{subfigure}[t]{0.25\columnwidth}
         \centering
         \includegraphics[width=1.8cm]{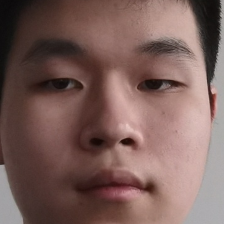}
         \caption*{$d$: $\{ S,O\}$}     
     \end{subfigure}%
     \hspace{1mm}
        \begin{subfigure}[t]{0.25\columnwidth}
         \centering
         \includegraphics[width=1.8cm]{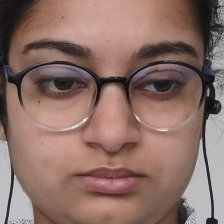}
         \caption*{$e$: $\{ G\}$}     
     \end{subfigure}%
     \hspace{1mm}
     \begin{subfigure}[t]{0.25\columnwidth}
         \centering
         \includegraphics[width=1.8cm]{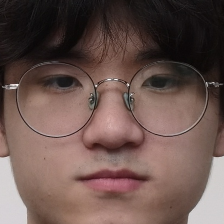}
         \caption*{$f$: $\{ G, O\}$}     
     \end{subfigure}%
    \vspace{0.1mm} 
     \begin{subfigure}[t]{0.25\columnwidth}
         \centering
         \includegraphics[width=1.8cm]{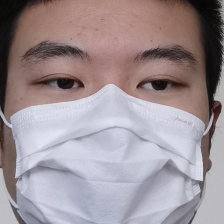}
         \caption*{$g$: $\{ M\}$}     
     \end{subfigure}%
     \hspace{1mm}
     \begin{subfigure}[t]{0.25\columnwidth}
         \centering
         \includegraphics[width=1.8cm]{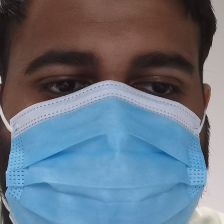}
         \caption*{$h$: $\{ M, O\}$}     
     \end{subfigure}%
     \hspace{1mm}
     \begin{subfigure}[t]{0.25\columnwidth}
         \centering
         \includegraphics[width=1.8cm]{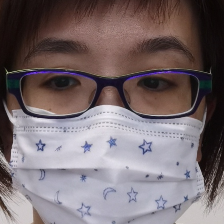}
         \caption*{$i$: $\{ G, M, O\}$}     
     \end{subfigure}%

    \caption{
    The nine session types in the RealGaze dataset.
    Sessions vary by illumination (light‑off $O$, side‑lit $S$) and accessories (glasses $G$, mask $M$), except Session $a$, which uses standard indoor lighting without accessories.
    }
     \label{fig:realgaze}
\end{figure}
\begin{table*}[tb]
  \caption{Comparative Evaluation on RealGaze (errors in mm).
  The best results are in \textbf{bold} and the second best results are with \underline{underline}.
  }
  \label{tab:alex2}
  \centering
  \resizebox{\textwidth}{!}{%
      \begin{tabular}{@{}l c | rrr | rrr | rrr | rrr | rrr@{}}
        \toprule
        \textbf{Model} (Backbone, Training data) & \multicolumn{1}{c}{\textbf{Model}} & \multicolumn{3}{c}{\textbf{Overall}} & \multicolumn{3}{c}{\textbf{Ideal}} & \multicolumn{3}{c}{\textbf{Side-Lit}} & \multicolumn{3}{c}{\textbf{Glasses}} & \multicolumn{3}{c}{\textbf{Masks}} \\
     & \multicolumn{1}{c}{\textbf{size}} & \multicolumn{1}{c}{$d_X$} & \multicolumn{1}{c}{$d_Y$} & \multicolumn{1}{c}{$\|d\|_2$} & \multicolumn{1}{c}{$d_X$} & \multicolumn{1}{c}{$d_Y$} & \multicolumn{1}{c}{$\|d\|_2$} & \multicolumn{1}{c}{$d_X$} & \multicolumn{1}{c}{$d_Y$} & \multicolumn{1}{c}{$\|d\|_2$} & \multicolumn{1}{c}{$d_X$} & \multicolumn{1}{c}{$d_Y$} & \multicolumn{1}{c}{$\|d\|_2$} & \multicolumn{1}{c}{$d_X$} & \multicolumn{1}{c}{$d_Y$} & \multicolumn{1}{c}{$\|d\|_2$} \\
        \midrule\midrule
    
        PureGaze (ResNet-50, $D_X$) & 31.0M & 46.6 & 71.0 & 93.5 & 35.3 & 73.3 & 85.7 & 59.4 & 54.3 & 90.3 & 56.1 & 79.6 & 106.0 & 31.7 & 76.7 & 88.4\\
        ETH-XGaze (ResNet-50, $D_X$) & 25.6M & 42.9 & 59.7 & 82.1 & 24.8 & 62.7 & 72.3 & 33.0 & 56.9 & 71.8 & 73.3 & 56.6 & 104.8 & 31.3 & 62.0 & 75.1\\
        UniGaze-B-Joint (ViT-B, 5 datasets) & 86.6M & 21.1 & 44.4 & 52.8 & 16.1 & 33.7 & 40.6 & \underline{17.4} & 35.0 & 41.8 & 26.9 & 52.9 & 63.8 & 19.7 & 40.2 & 48.6\\
        UniGaze-H-Joint (ViT-H, 5 datasets) & 632M & \textbf{18.3} & 44.1 & 51.5 & \textbf{13.9} & 38.1 & 43.1 & \textbf{14.6} & 35.5 & 41.3 & \underline{25.1} & 47.9 & 59.0 & \textbf{16.0} & 37.9 & \textbf{45.3}\\
        \textbf{Ours} (MobileNet-v2, $\{D_X, D_N, D_C\}$) & 3.8M & 22.3 & \textbf{34.9} & \underline{46.3} & 16.0 & \textbf{28.8} & \underline{36.6} & 18.6 & \underline{28.5} & \underline{37.0} & 25.4 & \textbf{33.4} & \underline{46.6} & 20.9 & \textbf{36.0} & \textbf{45.3}\\
        \textbf{Ours} (ViT-B, $\{D_X, D_N, D_C\}$) & 86.6M & \underline{19.1} & \underline{35.8} & \textbf{44.4} & \underline{14.9} & \underline{29.3} & \textbf{36.2} & 17.7 & \textbf{27.9} & \textbf{35.9} & \textbf{24.4} & \underline{33.5} & \textbf{46.0} & \underline{18.8} & \underline{36.5} & \textbf{45.3}\\
      \bottomrule
      \end{tabular}
    }
\end{table*}

\section{Benchmark Datasets}
\label{s:benchmark}

Cross‑dataset evaluation is the standard protocol for assessing the generalization ability of AGE models, yet existing test cases remain too narrow to capture the challenges of real‑world deployment. In particular, accessories such as eyeglasses and facial masks often cause severe performance degradation, but no standardized benchmark or metric currently isolates their impact. To expose the true robustness of AGE models in these long‑tail scenarios, we introduce two complementary benchmarks that provide fine‑grained measurements of degradation induced by wearables and environmental factors.

\subsection{RealGaze: A Real-World Benchmark}
\label{s:realgaze}
The RealGaze dataset emulates real-world application of AGE models on mobile devices.
The collection environment consists of a 13-inch tablet mounted on a fixed stand 0.5m from the subject.
We record 20 volunteers with diverse demographic attributes.
As illustrated in \cref{fig:realgaze}, each subject completes nine session types spanning combinations of wearables (none, glasses, mask) and illumination conditions (standard indoor lighting, low‑light, and harsh side‑lighting). Each session contains 100 paired samples of PoG locations and front‑camera images, providing a statistically robust basis for quantifying how specific environmental and personal factors degrade gaze estimation performance.

\subsection{ZeroGaze: A Controlled Synthetic Benchmark}
\label{s:zerogaze}
To isolate the impact of head pose and gaze variance from ocular appearance, we introduce ZeroGaze. Leveraging the Flux.1 \cite{flux} text-to-image model, we generate a massive-scale synthetic dataset consisting of 10,000 image triplets $\{X, X_g, X_m\}$, as shown in \cref{fig:teaser}(a).
In each triplet, $X$ is a clean face, while $X_g$ and $X_m$ feature the same identity and geometry, with glasses and masks being the only visual variables.
Through precision prompting and using facial-feature similarity scores computed with ArcFace\cite{arcface}, we ensure that all images are rendered with near‑zero gaze and head‑pose labels. 
As a result, any deviation from the zero‑degree ground truth directly reflects the model's inability to marginalize gaze-irrelevant features. ZeroGaze therefore provides a clean ``zero‑point'' calibration metric for quantifying wearable‑induced gaze bias. 


\section{Experiments}
\label{s:exp}

\subsection{Experimental Setup}
\label{s:setup}
We utilize ETH-XGaze $D_X$ \cite{xgaze}, GazeGene $D_N$ \cite{gazegene}, and GazeCapture $D_C$ \cite{gazecapture} for training.
To ensure domain alignment, we filter all datasets to a shared 
head pose interval $I_H: [-30^\circ, 30^\circ] \times [-30^\circ, 30^\circ]$ and a gaze interval $I: [-30^\circ, 14^\circ] \times [-26^\circ, 26^\circ]$.
This range covers the dominant interaction envelope for mobile and laptop use. 
$I$ is partitioned into $13 \times 11$ bins with a bin size of $s_\phi = s_\psi = 4^\circ$.
As motivated in \cref{s:joint}, we disable direct pitch supervision for $D_N$ and $D_C$.
Additional  details are provided in the Appendix.

We benchmark against several open-source models:
PureGaze \cite{puregaze}, ETH-XGaze ResNet-50 \cite{xgaze}, and the ViT-based UniGaze \cite{unigaze}, the current SOTA.
For fairness, all model outputs are clamped to the interval $I$.

\subsection{RealGaze Evaluation}
\label{s:exp_realgaze}

We evaluate generalization under real‑world conditions using the nine RealGaze session types (\cref{fig:realgaze}).
By averaging over specific subsets, we analyze five settings: \textbf{Overall} (all sessions), \textbf{Ideal} ($a$, $b$), \textbf{Side‑Lit ($c$, $d$)}, \textbf{Glasses ($e$, $f$)}, and \textbf{Masks ($g$, $h$)}.
We map the predicted 3D gaze vectors to 2D screen coordinates and report the average Euclidean error $\|d\|_2$, as well as error components along the X ($d_X$) and Y ($d_Y$) axes in millimeters.

As shown in \cref{tab:alex2}, our method outperforms the competition by a large margin, particularly in $d_Y$ and $\|d\|_2$.
The performance gap is most pronounced in the 
Glasses and Masks sessions:
while existing models suffer substantial degradation, our approach maintains stable performance by consistently attending to the ocular region rather than accessory‑induced artifacts.
We also observe a reduced accuracy loss in the Side-lit sessions in our models compared to the baselines.

\subsection{ZeroGaze Evaluation}
\label{s:exp_zerogaze}

\begin{figure*}[t]
     \centering
     \captionsetup[subfigure]{justification=centering, singlelinecheck=false}
     
     \begin{subfigure}[t]{0.5\textwidth}
         \centering
         \includegraphics[height=3.0cm]{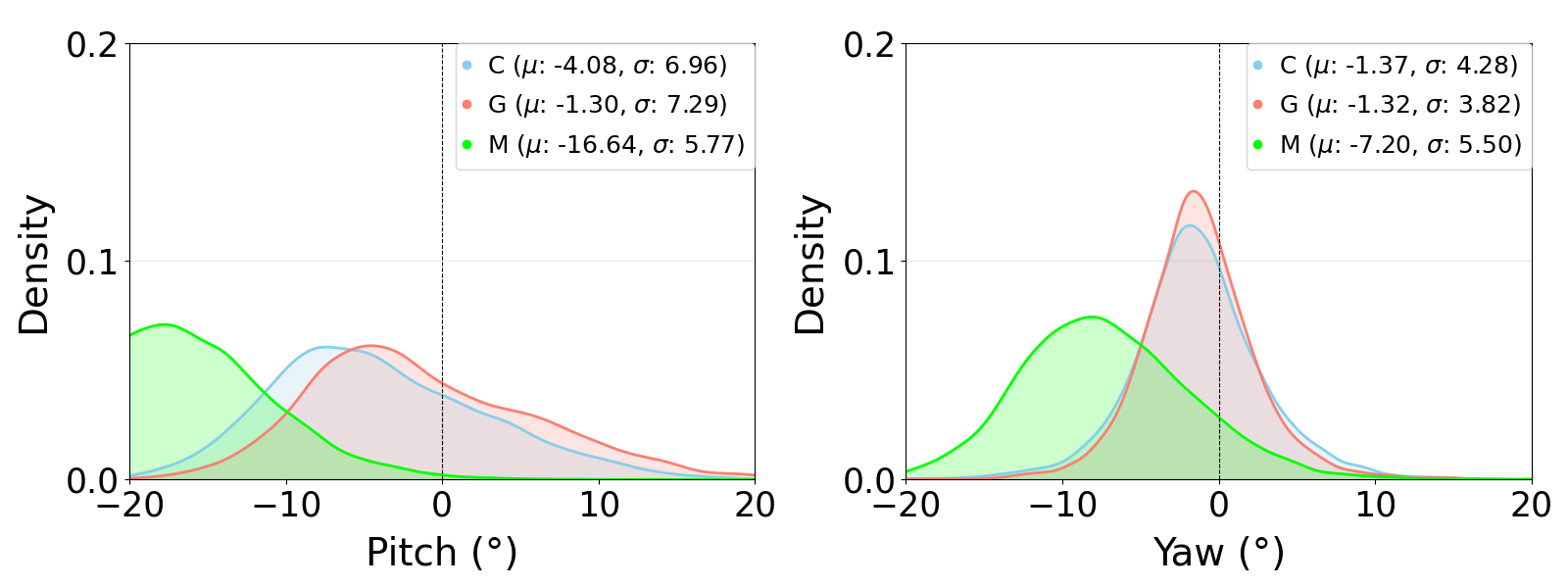}
         \caption{ETH-XGaze (ResNet-50)}
     \end{subfigure}%
     \begin{subfigure}[t]{0.5\textwidth}
         \centering
         \includegraphics[height=3.0cm]{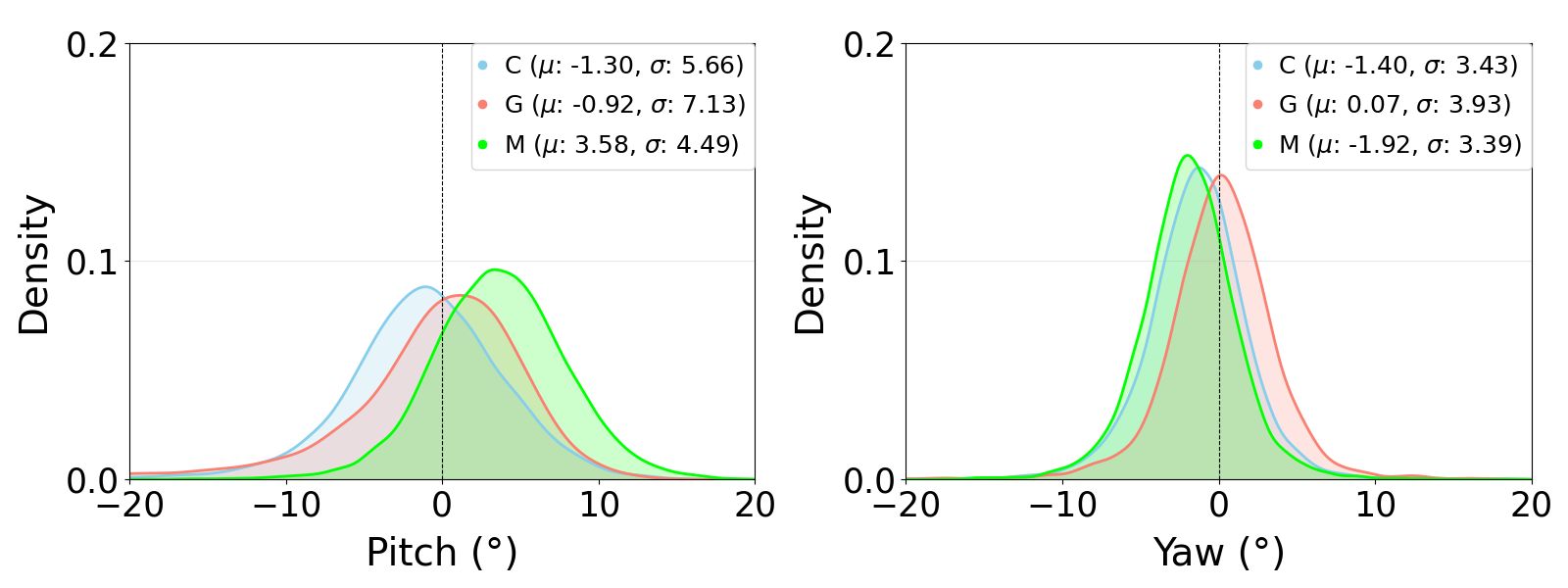}
         \caption{UniGaze-B-Joint (ViT-B)}
     \end{subfigure}


     \begin{subfigure}[t]{0.5\textwidth}
         \centering
         \includegraphics[height=3.0cm]{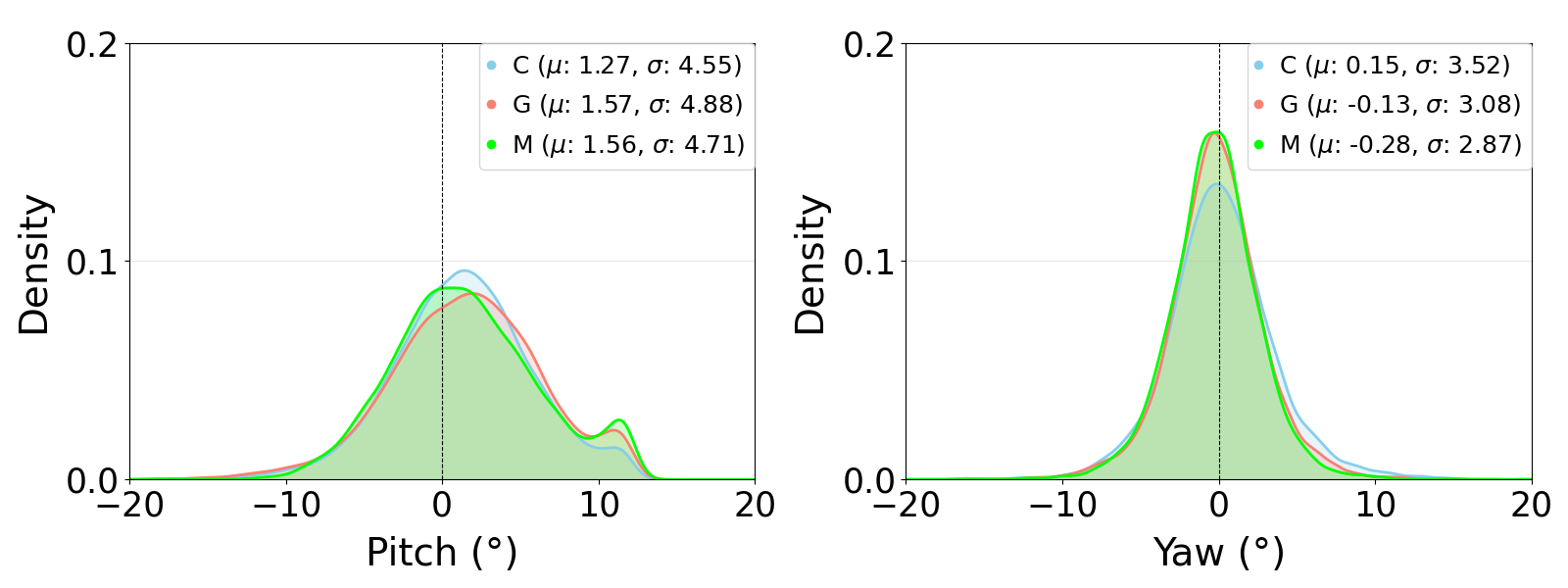}
         \caption{Ours (MobileNet)}
     \end{subfigure}%
     \begin{subfigure}[t]{0.5\textwidth}
         \centering
         \includegraphics[height=3.0cm]{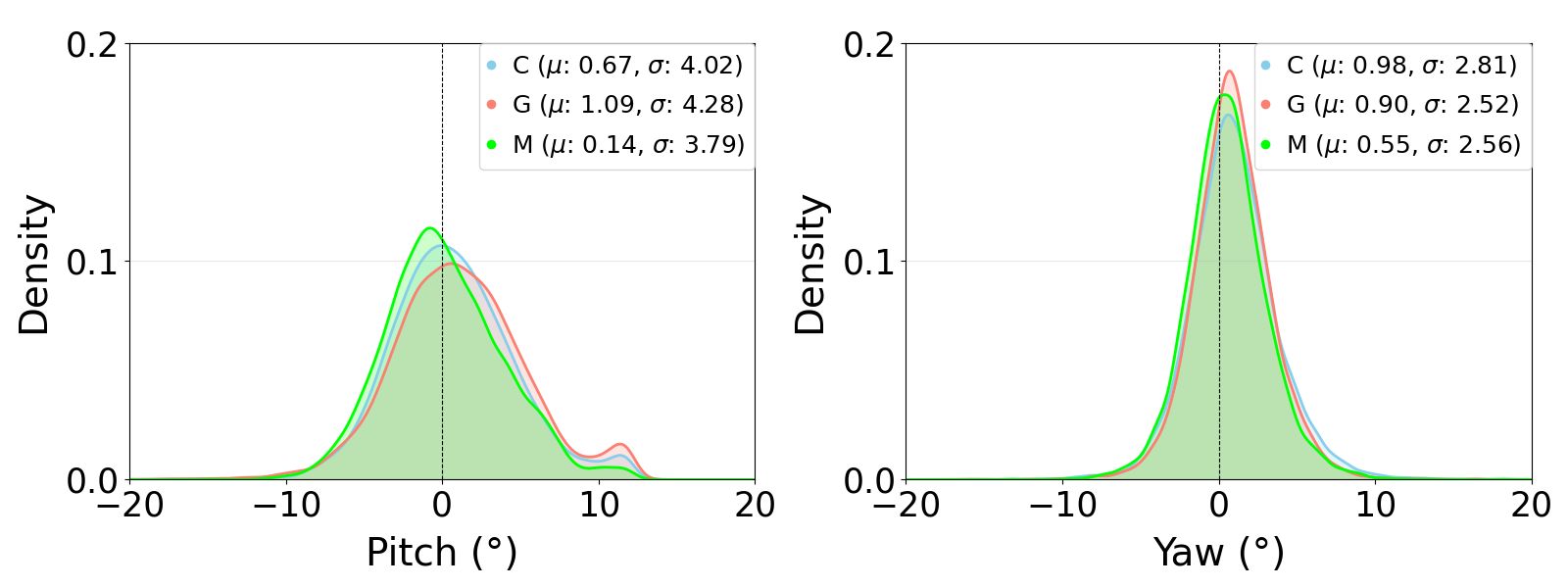}
         \caption{Ours (ViT-B)}
     \end{subfigure}

     \caption{
     Distribution of AGE results on ZeroGaze across three views: \textbf{C}lean (blue), \textbf{G}lasses (red), and \textbf{M}asks (green).
     Our methods maintain concentrated and zero-centered, while baselines yield biased results with long tails, particularly along pitch.
     }
     \label{fig:zerogaze_dist}
\end{figure*}
\begin{figure}[t] 
     \centering
     \captionsetup[subfigure]{justification=centering, singlelinecheck=false}
     \scriptsize 

     \begin{subfigure}[t]{0.4\columnwidth}
         \centering
         \includegraphics[width=2.5cm]{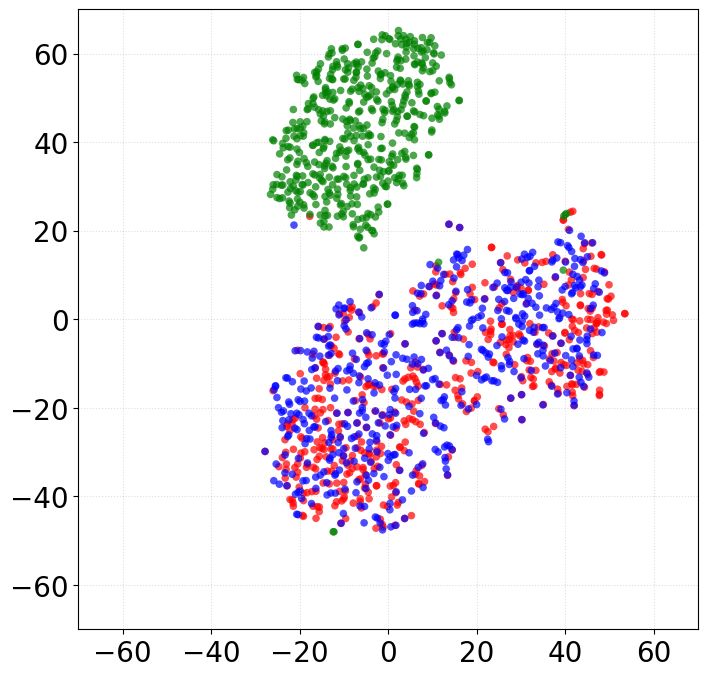}
         \caption{ETH-XGaze}
     \end{subfigure}%
     \hspace{1mm}
     \begin{subfigure}[t]{0.4\columnwidth}
         \centering
         \includegraphics[width=2.5cm]{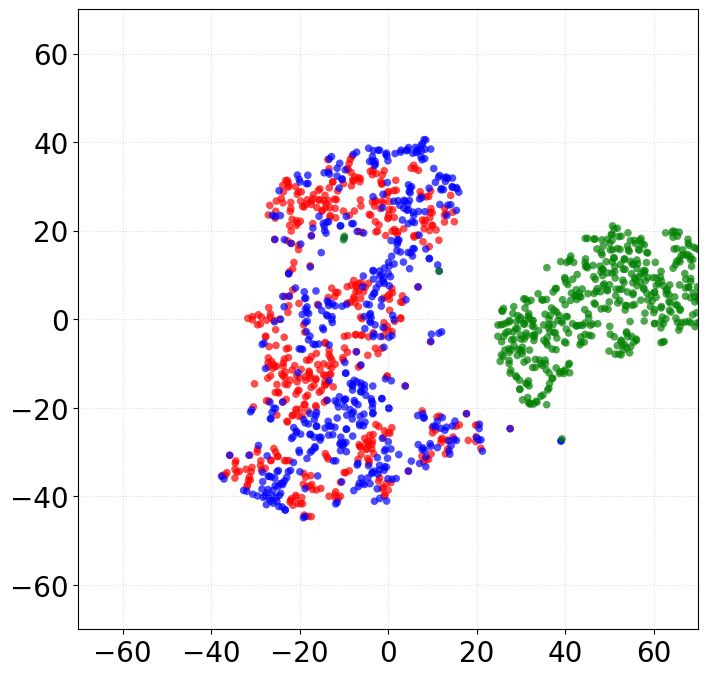}
         \caption{UniGaze-B-Joint}
     \end{subfigure}%
     
     \vspace{1mm} 

     \begin{subfigure}[t]{0.4\columnwidth}
         \centering
         \includegraphics[width=2.5cm]{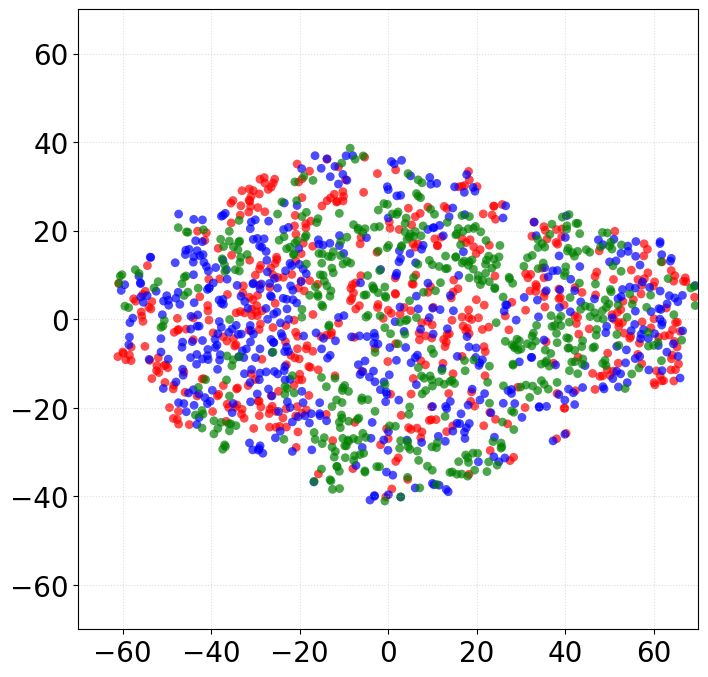}
         \caption{Ours (MobileNet)}
     \end{subfigure}%
     \hspace{1mm}
     \begin{subfigure}[t]{0.4\columnwidth}
         \centering
         \includegraphics[width=2.5cm]{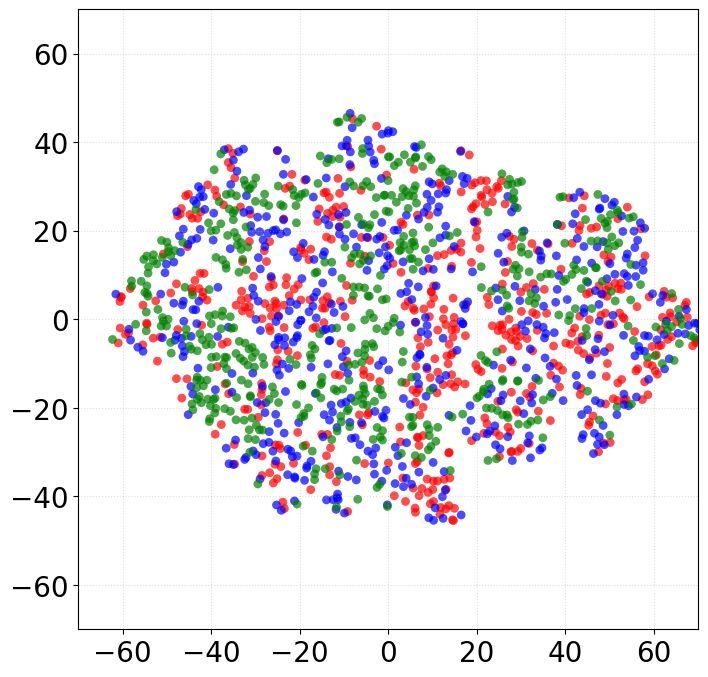}
         \caption{Ours (ViT-B)}
     \end{subfigure}
     
\caption{t-SNE visualization of high-level features on ZeroGaze.
Our models successfully learn occlusion-invariant feature manifolds across the three views: 
Clean (blue), Glasses (red) and Masks (green).
}

     \label{fig:zerogaze_tsne}
\end{figure}

Since all ZeroGaze samples possess a ground-truth label of zero, any non-zero prediction represents a purely visual bias introduced by facial features or accessories.

\cref{fig:zerogaze_dist} visualizes the output distributions on ZeroGaze. Existing models produce wide, off‑center distributions with heavy pitch‑axis tails, indicating that glasses and masks are frequently misinterpreted as vertical gaze shifts. Notably, the ETH‑XGaze model exhibits a persistent negative pitch bias, reflecting a collapse toward the mean gaze vector of $D_X$ rather than a faithful mapping of absolute gaze.
Our method, conversely, yields highly concentrated distributions around the zero-point.
Further analysis via t-SNE (\cref{fig:zerogaze_tsne}) confirms that our backbone learns an occlusion-invariant feature space, effectively ``seeing through'' the occlusions that confuse standard regressors.

\subsection{Ablation Studies}
\label{s:ablation}
We conduct a series of ablations on our MobileNet model to isolate the contribution of each design component.

\begin{figure*}[tb]
     \centering
     \captionsetup[subfigure]{justification=centering, singlelinecheck=false}
     \scriptsize 
     
     \begin{subfigure}[t]{1\textwidth}
         \centering
         \includegraphics[width=\textwidth]{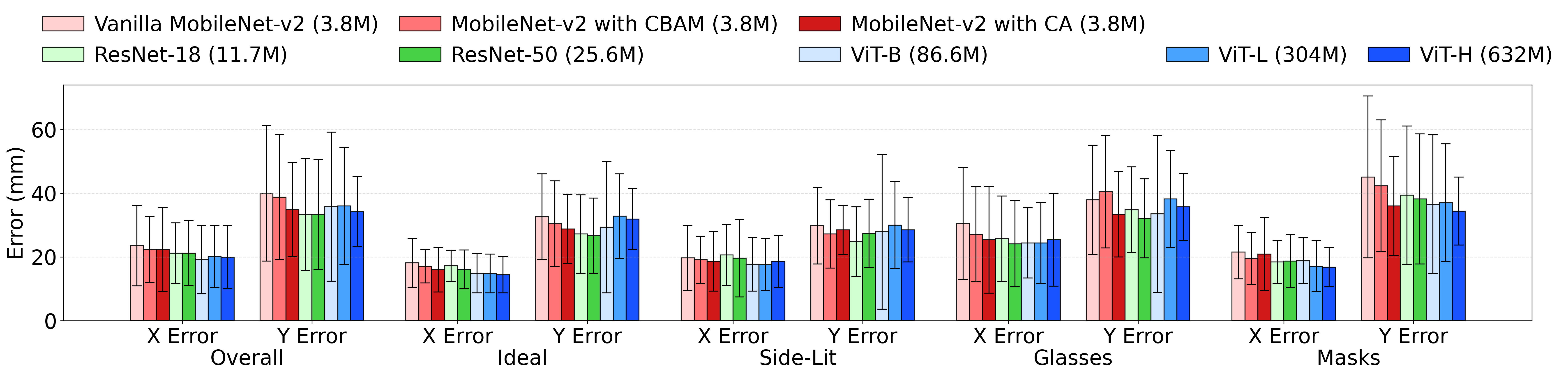}
     \end{subfigure}%
     
     \caption{
     Performance benchmark across various backbone architectures on RealGaze.
     }
     \label{fig:backbone}
\end{figure*}

\paragraph{Backbone}
As shown in \cref{fig:backbone}, lightweight backbones such as MobileNet benefit substantially from attention modules such as CBAM \cite{cbam} and CA \cite{ca}, which provide notable accuracy gains at negligible cost. As model capacity increases, performance saturates around the scale of ResNet‑50, indicating a nonlinear relationship between model size and accuracy. Scaling further to ViT‑level models marginally reduces variance and $d_X$, but yields diminishing returns in $d_Y$.

\begin{table*}[tb]
  \caption{Ablation studies on RealGaze (errors in mm).
  }
  \label{tab:alex2_ablation}
  \centering
  \resizebox{0.9\textwidth}{!}{%
      \begin{tabular}{@{}l rrr | rrr | rrr | rrr | rrr@{}}
        \toprule
        \textbf{Model} & \multicolumn{3}{c}{\textbf{Overall}} & \multicolumn{3}{c}{\textbf{Ideal}} & \multicolumn{3}{c}{\textbf{Side-Lit}} & \multicolumn{3}{c}{\textbf{Glasses}} & \multicolumn{3}{c}{\textbf{Masks}} \\
         & \multicolumn{1}{c}{$d_X$} & \multicolumn{1}{c}{$d_Y$} & \multicolumn{1}{c}{$\|d\|_2$} & \multicolumn{1}{c}{$d_X$} & \multicolumn{1}{c}{$d_Y$} & \multicolumn{1}{c}{$\|d\|_2$} & \multicolumn{1}{c}{$d_X$} & \multicolumn{1}{c}{$d_Y$} & \multicolumn{1}{c}{$\|d\|_2$} & \multicolumn{1}{c}{$d_X$} & \multicolumn{1}{c}{$d_Y$} & \multicolumn{1}{c}{$\|d\|_2$} & \multicolumn{1}{c}{$d_X$} & \multicolumn{1}{c}{$d_Y$} & \multicolumn{1}{c}{$\|d\|_2$} \\
        \midrule\midrule
        \textbf{Ours, baseline} (MobileNet) & 22.3 & 34.9 & 46.3 & 16.0 & 28.8 & 36.6 & 18.6 & 28.5 & 37.0 & 25.4 & 33.4 & 46.6 & 20.9 & 36.0 & 45.3\\
        \midrule
        (a) Training Data \\
        $D_X$ only & 32.9 & 45.4 & 61.9 & 23.5 & 41.8 & 52.0 & 31.0 & 34.8 & 52.1 & 36.5 & 43.6 & 63.0 & 35.3 & 52.0 & 69.2 \\
        $\{D_X, D_C\}$ & 26.1 & 45.1 & 57.2 & 18.8 & 47.4 & 54.5 & 23.3 & 31.5 & 43.2 & 31.6 & 45.3 & 61.1 & 25.7 & 56.2 & 66.9
 \\
        $\{D_X, D_C, D_{360}\}$ & 25.7 & 43.8 & 55.6 & 18.7 & 36.8 & 44.6 & 23.4 & 31.9 & 44.1 & 30.5 & 45.2 & 59.8 & 21.7 & 48.5 & 57.4 \\
        $\{D_X, D_C, D_N\}$ using all pitch labels & 22.1 & 51.4 & 60.1 & 17.2 & 48.3 & 54.6 & 19.8 & 42.1 & 50.6 & 26.1 & 59.7 & 70.5 & 19.1 & 44.8 & 52.4
  \\
        
        \midrule
        \multicolumn{15}{l}{(b) Classification Loss Term $L_{clf}$, and bin size $s_{\phi, \psi}$ (baseline $s=4$) } \\
        Without resampling or $L_{clf}$ & 30.0 & 46.9 & 61.0 & 21.5 & 35.9 & 46.0 & 25.0 & 33.8 & 46.4 & 37.0 & 41.9 & 61.3 & 28.9 & 47.8 & 61.4 \\
        Without $L_{clf}$ & 26.8 & 40.5 & 53.4 & 19.4 & 34.0 & 42.4 & 23.9 & 31.9 & 44.6 & 33.8 & 39.6 & 57.3 & 24.5 & 46.3 & 57.3 \\
        $s=1$ & 22.5 & 39.5 & 49.7 & 16.2 & 33.5 & 40.4 & 21.2 & 28.9 & 39.4 & 27.0 & 39.0 & 52.9 & 20.2 & 45.4 & 53.6\\
        $s=2$ & 25.2 & 41.6 & 53.3 & 20.0 & 34.1 & 43.0 & 23.0 & 29.1 & 41.1 & 29.6 & 42.4 & 57.3 & 23.9 & 48.9 & 59.1\\
        $s=8$ & 26.5 & 44.7 & 56.7 & 22.5 & 43.2 & 52.4 & 23.0 & 30.9 & 42.6 & 30.7 & 44.4 & 59.8 & 24.5 & 51.9 & 62.1\\
        \midrule
        (c) Without segmentation & 23.8 & 40.3 & 51.5 & 17.2 & 33.2 & 40.7 & 22.1 & 29.5 & 41.0 & 30.6 & 38.6 & 55.2 & 20.7 & 50.8 & 58.5
\\
        \midrule
        \multicolumn{15}{l}{(d) Augmentation and SupCon loss terms}\\
        Without augmentation & 29.0 & 56.9 & 69.4 & 19.0 & 50.8 & 58.1 & 24.3 & 34.3 & 46.2 & 39.0 & 59.6 & 78.2 & 21.6 & 58.2 & 66.2\\
         Without glasses synthesis & 24.8 & 39.3 & 51.4 & 17.4 & 32.1 & 39.8 & 23.0 & 32.0 & 43.5 & 35.1 & 39.3 & 58.9 & 19.1 & 45.3 & 53.2\\
         Without mask synthesis & 25.7 & 43.3 & 55.3 & 21.3 & 35.0 & 45.0 & 22.0 & 31.3 & 42.4 & 29.6 & 41.7 & 56.0 & 25.0 & 49.6 & 61.0\\
         
         Without $L_\phi^S$ & 22.7 & 41.5 & 51.9 & 16.4 & 41.8 & 48.4 & 20.7 & 36.5 & 44.7 & 28.6 & 41.6 & 56.5 & 20.3 & 47.1 & 55.6
 \\
         Without $L^S_D$ & 23.4  & 39.7  & 50.6  & 17.5  & 33.5  & 40.9  & 21.1  & 30.1  & 40.6  & 26.2  & 33.9  & 47.6  & 23.7  & 46.4  & 57.1
\\
         Without $L^S_g, L^S_m$ & 22.4 & 40.0 & 49.8 & 14.7 & 33.0 & 38.7 & 19.8 & 32.5 & 41.7 & 27.5 & 41.1 & 54.0 & 21.1 & 42.8 & 51.2 \\
      \bottomrule
      \end{tabular}
    }
\end{table*}

\paragraph{Training Data}
\cref{tab:alex2_ablation}(a) validates our hypothesis regarding inter-dataset label deviation. As training data increases, $d_X$ improves consistently, but $d_Y$ often worsens, especially in the Glasses sessions, due to inconsistent pitch labels across datasets.
By a softer SupCon loss rather than direct regression loss, we effectively insulate the learning process from
this label ``contamination,'' stabilizing vertical gaze estimation.
In addition, replacing $D_{N}$ with $D_{360}$ results in larger errors in all sessions.

\paragraph{Classification and Segmentation}
As shown in \cref{tab:alex2_ablation}(b), compared to the na\"{i}ve $\ell_1$-only baseline, resampling and discretized classification significantly improve accuracy on both axes.
We also show that $4^\circ$ bin size offers the best trade‑off between discretization error and classification stability.
The segmentation task further improves accuracy in occlusion‑heavy sessions by anchoring the model's attention to the ocular region (\cref{tab:alex2_ablation}(c)), ensuring that gaze‑irrelevant textures do not dominate the representation.
\begin{figure}[t]
     \centering
     \captionsetup[subfigure]{justification=centering, singlelinecheck=false}

     \begin{subfigure}[t]{0.4\columnwidth}
         \centering
         \includegraphics[width=3.0cm]{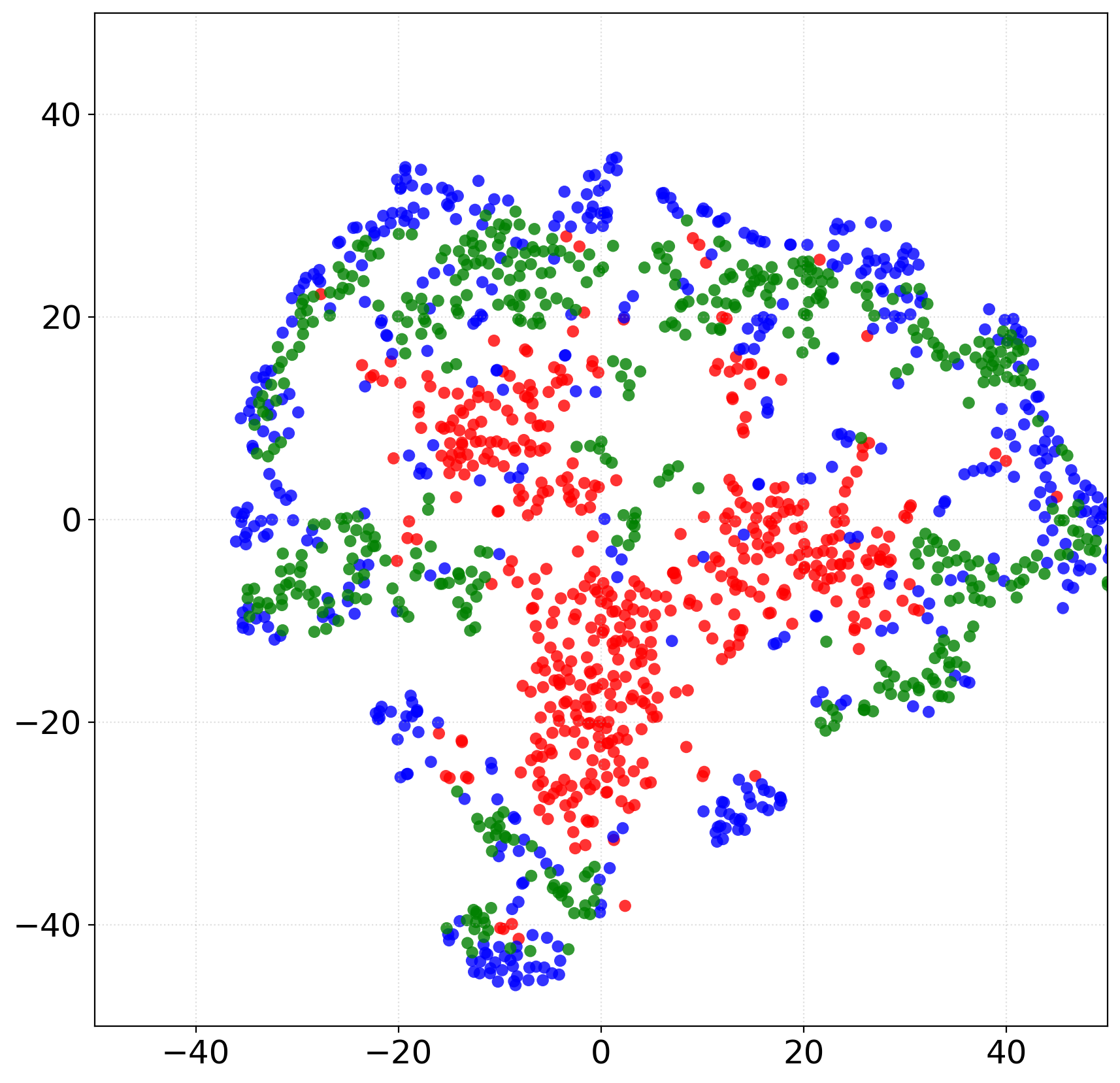}
         \caption{Without $L^{S}_{D}$}
     \end{subfigure}%
     \hspace{1mm}
     \begin{subfigure}[t]{0.4\columnwidth}
         \centering
         \includegraphics[width=3.0cm]{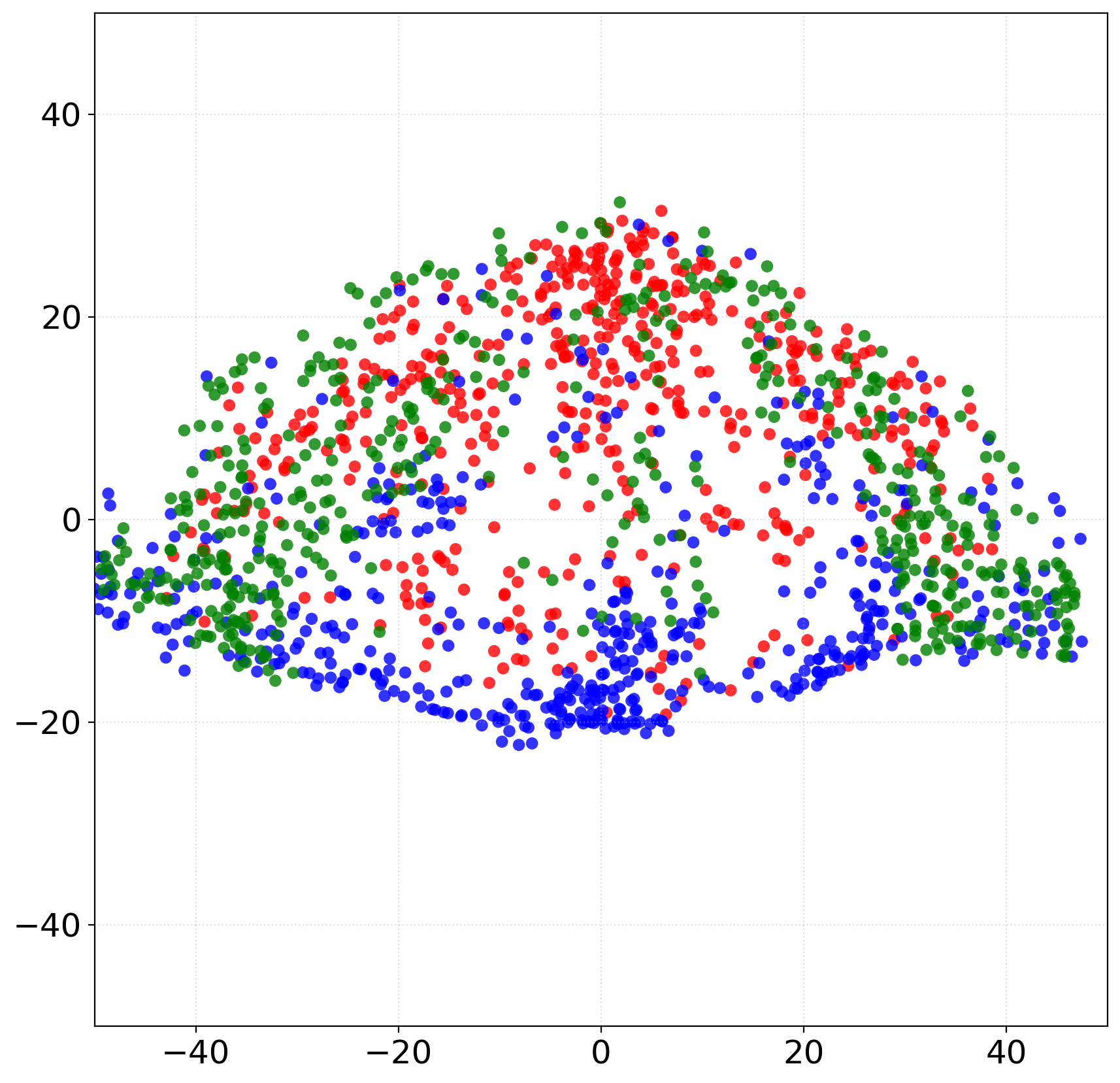}
         \caption{With $L^{S}_{D}$}
     \end{subfigure}

\caption{
Ablation of the dataset SupCon loss $L_D^S$.
t-SNE visualizations of features from $D_X$ (blue), $D_C$ (red), and $D_N$ (green) show that $L_D^S$ promotes a source-agnostic feature distribution, effectively marginalizing inter-dataset domain gaps.
}
\label{fig:tsne_dataset}
\end{figure}

\paragraph{Augmentation and SupCon}
Data augmentation is central to our generalization performance, as evidenced by the direct gains shown in \cref{tab:alex2_ablation}(d).
Remarkably, synthesizing glasses and masks alone improves robustness even in sessions without those accessories, indicating strong cross‑factor generalization.

All SupCon objectives contribute to consistent improvements in $d_Y$, with $L_{\phi }^S$ providing the largest benefit by stabilizing the pitch manifold. Although $L_D^S$ yields the smallest numerical gain, its structural role is essential: as illustrated in \cref{fig:tsne_dataset}, it produces a more cohesive, source‑agnostic feature distribution, demonstrating its effectiveness in marginalizing dataset‑specific signatures.

\section{Conclusion}

We identify limited image diversity and anisotropic inter‑dataset label deviation as key factors underlying the poor generalization of existing AGE models.
To counter them, we introduce a comprehensive framework that combines an automated augmentation pipeline with a multi‑task learning formulation. Evaluations on our RealGaze and ZeroGaze benchmarks demonstrate substantial gains in robustness and cross‑domain performance. Using MobileNet as the backbone, we deliver a lightweight yet effective solution suitable for real‑time deployment on mobile devices.

{
    \small
    \bibliographystyle{ieeenat_fullname}
    \bibliography{refs}
}

\clearpage
\maketitlesupplementary

\setcounter{section}{0}
\renewcommand{\thesection}{\Alph{section}}

\section{Implementation Details}

Models are trained on 8 nVidia V100 GPUs using
the Adam optimizer \cite{adam}. For the SupCon objectives, the temperature parameters $\tau_S$ is set to $0.07$.

For MobileNet-based model training, we initialize the backbone with ImageNet-pretrained weights and train for 50 epochs. The  batch size is 160 per GPU with a learning rate of $3 \times 10^{-3}$. Each epoch consists of 91,520 samples (640 per bin) drawn from each dataset. The weights for each loss term are: $\lambda_{clf} = 0.1$, $\lambda_{seg}= 0.05$, $\lambda^S_{\phi} = 0.005$, and $\lambda^S_{D} = \lambda^S_{g} = \lambda^S_{m} = 0.0025$.

For ViT-based model training, we use the pretrained weights from the encoder part of Masked Autoencoders (MAE) \cite{mae}, and reduce $\lambda^S_\phi$ to $0.0025$ and the learning rate to $10^{-3}$ to facilitate stable convergence. Due to the faster convergence rate observed than MobileNet, these models are trained for 15 epochs with a batch size of 96 per GPU.

\paragraph{Data processing for GazeCapture $G_C$}
\label{s:imp_gc3d}

As $G_C$ lacks official 3D gaze labels, we adopt the labels generated by FAZE \cite{faze}. $G_C$ presents significant quality and balance challenges due to its crowdsourced nature,
with subject sample counts varying between 18 and 3,529, and also high variance in image quality, including motion blur and eye closure.
To ensure a robust training signal,we discard samples where face detection fails or where eye blinking is detected via eye landmarks.
During the stratified resampling for each bin (\cref{s:joint} in the main paper),
we perform subject-ID frequency balancing as well, to prevent overfitting to the subjects with dominant sample count.

\section{Further Discussion on Data Constraints}

\subsection{On Gaze Label Fidelity}

\begin{figure*}[tb]
  \centering
  \includegraphics[width=0.95\linewidth]{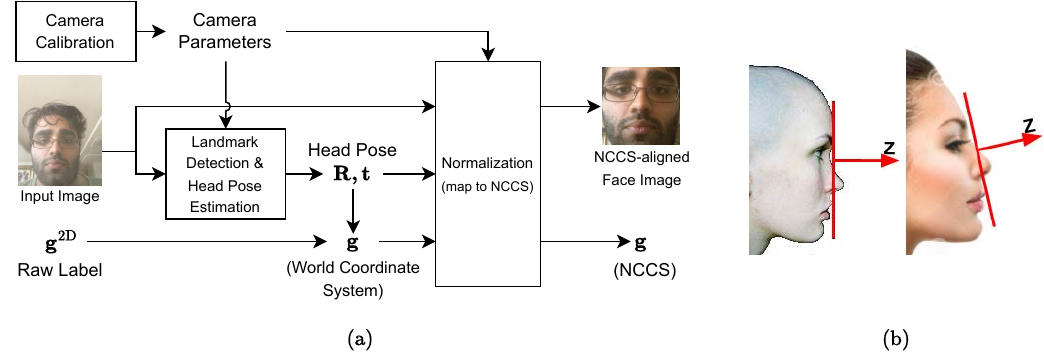}
  \caption{
  (a) Standard normalization pipeline for acquiring NCCS-aligned face images and corresponding 3D gaze vectors $\mathbf{g}$.
  (b) Morphological variance in skull structures induces ambiguity in the individual head coordinate systems, particularly along the vertical axis
  (image from \cite{differential}).
  }
  \label{fig:data_pipeline}
\end{figure*}

In most AGE datasets, the ground truth is established by recording a gaze target on a screen, represented by a 2D coordinate $\mathbf{g}^{2D}$.
$\mathbf{g}^{2D}$ is subsequently mapped to a 3D gaze vector $\mathbf{g}$ within the normalized camera coordinate system (NCCS) via a normalization operator \cite{normalization}.
This normalization process crops and warps the face image to simulate a front-facing subject located at a fixed distance along the camera's optical axis.
This alignment is highly beneficial as it eliminates several confounding variables, such as varying subject-to-camera distances and head-pose roll angles,
thereby stabilizing the input to the AGE model. 
However, this pipeline is vulnerable to accumulated errors from multiple stochastic and systemic sources:
\begin{itemize}
    \item \textbf{Angle Kappa}, which is the physiological pattern of human eye in which its visual axis is not aligned with its optical axis, introduces a person-specific error of typically 2 to 3 degrees \cite{kappa}.
    It is present in nearly all gaze datasets.
    \item \textbf{Error in Raw Label ($\mathbf{g}^{2D}$)},
    which arises from data acquisition lag or subject distraction, is frequently seen in crowdsourced datasets like GazeCapture \cite{gazecapture}.
    It propagates through the normalization pipeline,
    resulting in noisy 3D labels.
    \item \textbf{6D Head Pose Estimation (HPE)}.
    The pipeline estimates 6D head pose (position and orientation) by matching 2D facial landmarks to a canonical 3D head model. Discrepancies between an individual's unique skull structure and the average head model, combined with ambiguities in defining ``zero'' head pose (\cref{fig:data_pipeline}(b)), lead to distorted face images and erroneous $\mathbf{g}$.
    \item The \textbf{Camera Intrinsic Parameters}.
    The mapping between the 2D image and the 3D world coordinate system, along with HPE, rely on the camera's intrinsic matrix $\mathbf{k}$ and sensor size $(w_c, h_c)$.
    These parameters are frequently unknown and instead estimated by camera calibration, leading to global 3D label deviation.
    \item \textbf{Discrepancy in Coordinate Systems}. Datasets that do not utilize the NCCS pipeline (\eg, $D_{360}$ and $D_N$) have a fundamental domain gap 
    in both image appearance and label distribution to those NCCS-aligned.
\end{itemize}
As a result, the interplay of these factors creates the inter-dataset label deviation that our proposed framework seeks to marginalize.

\subsection{Why is inter-dataset label deviation anisotropic?}

\begin{figure}[t]
    \centering
    \includegraphics[width=0.65\columnwidth]{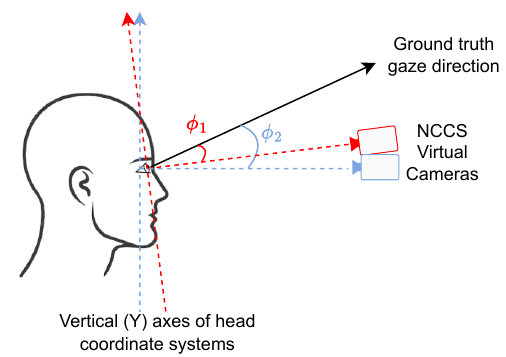}
    \caption{Divergent HPE results lead to a systemic deviation between the resulting NCCS pitch gaze labels $\phi_1$ and $\phi_2$.}
    \label{fig:nccs}
\end{figure}

Among the aforementioned factors,
we identify HPE as the main culprit of anisotropic label noise within the NCCS pipeline.
The spirit of the normalization operator is to ``relocate'' the camera so that the resulting 
NCCS satisfies two geometric principles:
1) its X-axis aligns with the X-axis of the head coordinate system (HCS) with respect to the subject; and
2) its Z-axis passes through the midpoint of the eyes (the origin of gaze), with the 
eyes maintained at a constant distance from the NCCS origin.
As shown in \cref{fig:nccs}, the definition of NCCS is not unique but rather hinges on the result of HPE.
While the bilateral symmetry of human face facilitates relatively reliable estimation of yaw and roll components,
morphological variations, specifically the diversity in jawlines, hinder the precise estimation of the pitch angle. This creates a systemic deviation in pitch gaze labels $\phi$.
Furthermore, the image warping performed by the normalization is a perspective transformation; although this transformation can simulate a nudge in head pose, it cannot synthesize the complex changes in ocular appearance that would naturally occur with a true change in $\phi$.
Consequently, in the case of \cref{fig:nccs}, 
two divergent HPE results can
yield a pitch label deviation of $|\phi_2 - \phi_1|$ with the warped face images not exhibiting a corresponding shift in gaze sense.
This appearance-label mismatch is the root cause of inter-dataset pitch label deviation.

\subsection{Analysis of Label Fidelity in Existing Datasets}
\label{s:app_datasets}

As detailed in \cref{tab:app_dataset}, we investigate the image quality along with label fidelity of six frequently used datasets in recent AGE works,
by examining systemic vulnerabilities to the factors identified above (excluding the ubiquitous angle kappa).
\begin{table*}[t]
  \caption{
  Comparative audit of dataset scale and 3D gaze label fidelity.
  Overall ratings assess the systematic reliability of labels within the NCCS framework across three tiers: \textbf{High}: The label is consistently precise with minimal documented bias;
  \textbf{Medium}: The label is generally acceptable for training but subject to localized systemic errors (e.g., hardware-limited HPE or distribution skew);
  \textbf{Low}: The label is considered suspicious, suffering from severe systemic noise originating from multiple sources such as non-standardized coordinate systems or extreme environmental constraints.
  }
  \label{tab:app_dataset}
  \centering
  \resizebox{\textwidth}{!}{%
  \begin{tabular}{@{}lccccccc@{}}
    \toprule
    Dataset & \makecell{Subject \#} & \makecell{Sample \#} & \makecell{Cam. Param. Fidelity} & \makecell{Raw Label Fidelity } & \makecell{Image Quality \& HPE Fidelity} & NCCS & \makecell{Overall Fidelity} \\
    \midrule
    $D_E$: EYEDIAP (FT sessions excluded) \cite{eyediap} & 14 & 15K & Medium & High & Low & Yes & Low \\
    $D_C$: GazeCapture \cite{gazecapture} & 1474 & 2M & Medium & Medium & Medium & Yes & Medium \\
    $D_M$: MPIIFaceGaze \cite{mpiiface} & 15 & 45K & Medium & High & Medium &  Yes & Medium \\
    $D_3$: Gaze360 \cite{gaze360} & 238 & 172K & High & High & Low & No & Low \\
    $D_X$: ETH-XGaze \cite{xgaze} & 110 & 760K & High & High & High & Yes & High  \\ 
    $D_N$: GazeGene \cite{gazegene} & 56 & 1M & N/A & N/A & N/A  & No & Medium \\
  \bottomrule
  \end{tabular}
  }
\end{table*}
\begin{figure}[t] 
     \centering
     \captionsetup[subfigure]{justification=centering, singlelinecheck=false}
     \scriptsize 
      \begin{subfigure}[t]{0.25\columnwidth}
         \centering
         \includegraphics[width=\textwidth]{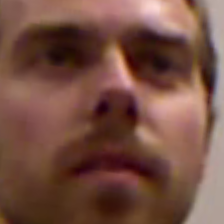}
         \caption{$D_E$} 
     \end{subfigure}%
      \hspace{1mm}
     \begin{subfigure}[t]{0.25\columnwidth}
         \centering
         \includegraphics[width=\textwidth]{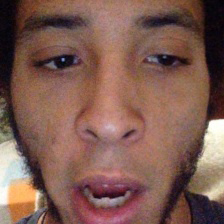}
         \caption{$D_C$} 
     \end{subfigure}%
      \hspace{1mm}
     \begin{subfigure}[t]{0.25\columnwidth}
         \centering
         \includegraphics[width=\textwidth]{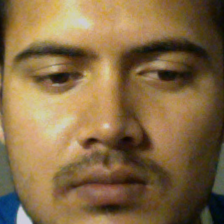}
         \caption{$D_M$} 
     \end{subfigure}%
      \vspace{0.1mm}
     \begin{subfigure}[t]{0.25\columnwidth}
         \centering
         \includegraphics[width=\textwidth]{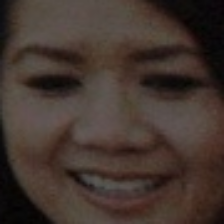}
         \caption{$D_{360}$} 
     \end{subfigure}%
      \hspace{1mm}
     \begin{subfigure}[t]{0.25\columnwidth}
         \centering
         \includegraphics[width=\textwidth]{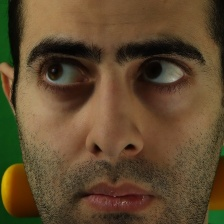}
         \caption{$D_X$} 
     \end{subfigure}%
      \hspace{1mm}
     \begin{subfigure}[t]{0.25\columnwidth}
         \centering
         \includegraphics[width=\textwidth]{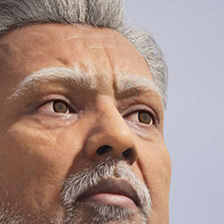}
         \caption{$D_N$} 
     \end{subfigure}%
     \caption{
     Representative samples of EYEDIAP $D_E$, GazeCapture $D_C$, MPIIFaceGaze $D_M$, Gaze360 $D_{360}$, ETH-XGaze $D_X$, and GazeGene $D_N$.
     }
     \label{fig:appendix_datasets}
\end{figure}

\paragraph{EYEDIAP} $D_E$ \cite{eyediap}
is the earliest dataset among the six. The raw data consists of 
video sequences paired with frame-wise labels for head pose, 2D screen coordinates ($\mathbf{g}^{2D}$), and camera parameters.
Our analysis follows the standard protocol by focusing exclusively on Continuous screen target (CS) and Discrete screen target (DS) sessions, with 14 human subjects involved.

$D_E$'s image quality is poor under the current standard, with a low frame resolution of $640 \times 480$. The extracted face patches must be upsampled by over $2\times$ to meet the model's $224\times 224$ input requirement, further introducing blur.

While $\mathbf{g}^{2D}$ labels are relatively reliable due to the controlled laboratory setting, the 3D head pose labels are problematic. They are generated using a Microsoft Kinect v1 depth sensor, with a native depth resolution of only $320\times 240$.
Previous investigations into the more advanced Kinect v2 reveals average pitch estimation errors exceeding $7^\circ$ \cite{kinectv2}; it is reasonable to assume the v1-derived labels in $D_E$ harbor even greater inaccuracies, particularly along the pitch axis.
Because $D_E$ lacks native 3D NCCS labels, researchers must rely on post-processing toolboxes \cite{recurrentgaze, gazehub} to derive gaze vectors. We find that these toolboxes produce inconsistent labels, with an average pitch label deviation of $5.7^\circ$. To conclude, $D_E$'s label fidelity is not among the best.

We also note that $D_E$ has a severely biased distribution, with $98\%$ of pitch labels concentrated in the narrow range of $[5^\circ, 21^\circ]$.
This is a direct artifact of the collection setup,
where the camera is 
positioned greatly below the screen,
resulting in almost all sample images depicting an upward gaze.
This distribution is antithetical to most deployment scenarios (e.g., smartphones or laptops), where cameras are positioned above the display, typically resulting in a slightly downward pitch.

\paragraph{GazeCapture} $D_C$ \cite{gazecapture}
With over 2 million samples and 1,450 participants, $D_C$ represents the most extensive demographic and environmental diversity available in the AGE domain. While the native image resolution is generally acceptable, the crowdsourced nature of the dataset introduces non-negligible label fidelity issues. 
First, $D_C$ is saturated with invalid samples, including those featuring severe motion blur, extreme under/over-exposure, and involuntary eye blinking.
Second, a noticeable portion of the raw $\mathbf{g}^{2D}$ labels exhibit low fidelity. The label errors stem from subject distraction (looking away from the target) and/or system latency in unconstrained mobile environments, where the data acquisition lag results in a mismatch between the captured image and the logged coordinate.
Third, there is a stark imbalance in both sample volume (as mentioned in \cref{s:imp_gc3d}) and the range of gaze labels captured per participant. This distribution forms a high risk of model over-fitting to ``dominant'' subjects.

Similar to $D_E$, $D_C$ only provides 2D labels, head pose, and estimated camera parameters. The reliance on third-party toolboxes to derive 3D NCCS labels, combined with the bottlenecks in raw label accuracy, results in a dataset where fidelity is sacrificed for scale.

\paragraph{MPIIFaceGaze} $D_M$ \cite{mpiiface}
often serves as a benchmark with small-scale, laboratory-controlled data, featuring 15 subjects with 3,000 samples each.
While its laboratory setting ensures a reliable set of 2D raw labels, the transition to 3D NCCS coordinates introduces uncertainty, due to errors from the approximated camera parameters and head pose angles.

\paragraph{Gaze360} $D_{360}$ \cite{gaze360}
is designed to maximize diversity across subjects, environmental settings, and the range of gaze and head poses.
However, this expansion in scope comes at the cost of severe degradation in image and label fidelity.
The data collection is particularly set 
where the subjects are positioned at an average distance of 2.2m from the camera, significantly further than in most AGE datasets. This results in face patches with extremely low effective resolution, characterized by sensor noise and severe radial distortion (\cref{fig:appendix_datasets}(d)).
Moreover, $D_{360}$ does not adopt the NCCS pipeline, producing a domain gap when compared to NCCS-compliant datasets.
This misalignment explains the poor inter-dataset label consistency results in \cref{s:labels}, and the consistently low accuracy observed when $D_{360}$ serves as the target domain in cross-dataset generalization studies \cite{xgaze, gazegene}. 

\paragraph{ETH-XGaze} $D_X$ \cite{xgaze} achieves a significant scale without compromising data quality through a 
laboratory environment. The acquisition setup consists of 18 synchronized DSLR cameras with varied orientations and a specialized seat with a head rest to stabilize the subject's position and head pose.
In this way, each time of data capture produces 18 concurrent high-resolution images,
representing a single PoG with 18 distinct head poses.
This rigorous setup minimizes stochastic noise in the NCCS pipeline, producing what is currently the most reliable NCCS-aligned data in the literature, while the inherent ambiguities in HPE (particularly along the pitch axis) still persist.
A limitation of $D_X$ is that its head pose distribution is 
concentrated around the 18 discrete camera locations rather than spanning the label space uniformly.
This poses a risk of introducing an inductive bias, where models potentially over-fit to these specific camera orientations rather than learning a truly continuous head-pose-to-gaze mapping.

\paragraph{GazeGene} $D_N$ \cite{gazegene} is a contemporary synthetic dataset generated via Unreal Engine.
The primary merit of synthetic data lies in the availability of abundant and precise labels derived from the underlying simulation geometry, allowing $D_N$ to provide an unprecedented variety of auxiliary ground truths.
However, $D_N$ presents two major drawbacks in the context of AGE.
First, the images lack the stochastic variance and photorealistic complexity required for robust open-domain generalization. They often look ``too perfect'', lacking camera sensor noise and natural specular reflections on the cornea (see \cref{fig:appendix_datasets}(f)), which can lead to domain-specific feature collapse.
Second, similar to $D_{360}$, the label generation pipeline of $D_N$ does not inherently align with the NCCS protocol, resulting in a disparity in both image appearance and label distribution.
Consequently, we categorize its overall label fidelity as ``medium'' and treat it as a source of diverse geometric signals rather than definitive ground truths.

\section{Data Augmentation Pipeline Details}

\subsection{Stochastic Mixing Protocol}

To ensure the model generalizes across a wide image manifold, we apply a randomized combination of augmentations for each input sample. Each augmentation method is applied according to the following probability distribution:
\begin{itemize}
    \item Color jitter: $p=1$,
    \item Background replacement: $p=0.95$,
    \item Illumination perturbation: $p=0.5$,
    \item Sensor noise: $p=0.5$,
    \item Glasses synthesis: $p=0.5$,
    \item Mask occlusion: $p=0.5$,
    \item Blur: $p=0.25$,
    \item Desaturation: $p=0.1$.
\end{itemize}

For details regarding horizontal flipping, please refer to ``Multi-view Supcon Learning'' section (\cref{s:joint}) in the main paper.

\subsection{Realistic Sensor Noise Modeling}


\begin{algorithm}
\caption{Sensor Noise Injection}
\label{alg:ycrcb_noise}
    \begin{algorithmic}[1]
    \Require RGB image $I \in [0, 1]^{224 \times 224 \times 3}$, luma noise strength $\alpha_Y=11$, chromatic noise strength $\alpha_C=15$, blotch size $b_C=2$
    \State $I_{YCrCb} \gets \text{RGB2YCrCb}(I)$
    \State $Y, Cr, Cb \gets \text{Split}(I_{YCrCb})$
    \If{$\alpha_Y > 0$}
        \State Generate $N_Y \sim \mathcal{N}(\mathbf{0}, \mathbf{I})^{224 \times 224}$
        \State $Y \gets Y + \alpha_Y N_Y$
    \EndIf
    \If{$\alpha_C > 0$}
        \State Generate $N_{C_r} \sim \mathcal{N}(\mathbf{0}, \mathbf{I})^{224 \times 224}$
        \State $N_{C_r} \gets \text{GaussianBlur}(N_{C_r}, \sigma=b_C)$ 
        \State $N_{C_r} \gets N_{C_r} / \text{std}(N_{C_r})$ \Comment{Normalize the field}
        \State $Cr \gets Cr + \alpha_C N_{C_r}$
        \State Repeat Line 8-11 for $Cb$
    \EndIf
    \State Clamp $Y, Cr, Cb$ to $[0,1]$
    \State $I' \gets \text{YCrCb2RGB}(\text{Merge}(Y, Cr, Cb))$
    \State \Return $I'$
    \end{algorithmic}
\end{algorithm}

We introduce a compact noise model designed to approximate the 
stochastic interference inherent in camera sensors.
As detailed in \cref{alg:ycrcb_noise}, 
we perform noise injection in the YCbCr color space to decouple luminance (brightness) from chrominance (color) components. In particular, we inject mottled, ``blotch-like'' noise into the Cb and Cr channels, to simulate the noise artifact often seen in low-light photos.

\subsection{Pose-Consistent Eyeglasses Synthesis} 

\begin{figure}[t]
     \centering
     \captionsetup[subfigure]{justification=centering, singlelinecheck=false}
     \scriptsize 
     
     \begin{subfigure}[t]{0.3\columnwidth}
         \centering
         \includegraphics[height=2.5cm]{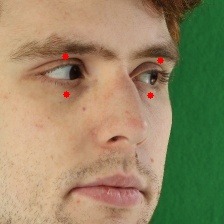}
     \end{subfigure}%
     \hspace{1mm}
     \begin{subfigure}[t]{0.3\columnwidth}
         \centering
         \includegraphics[height=2.5cm]{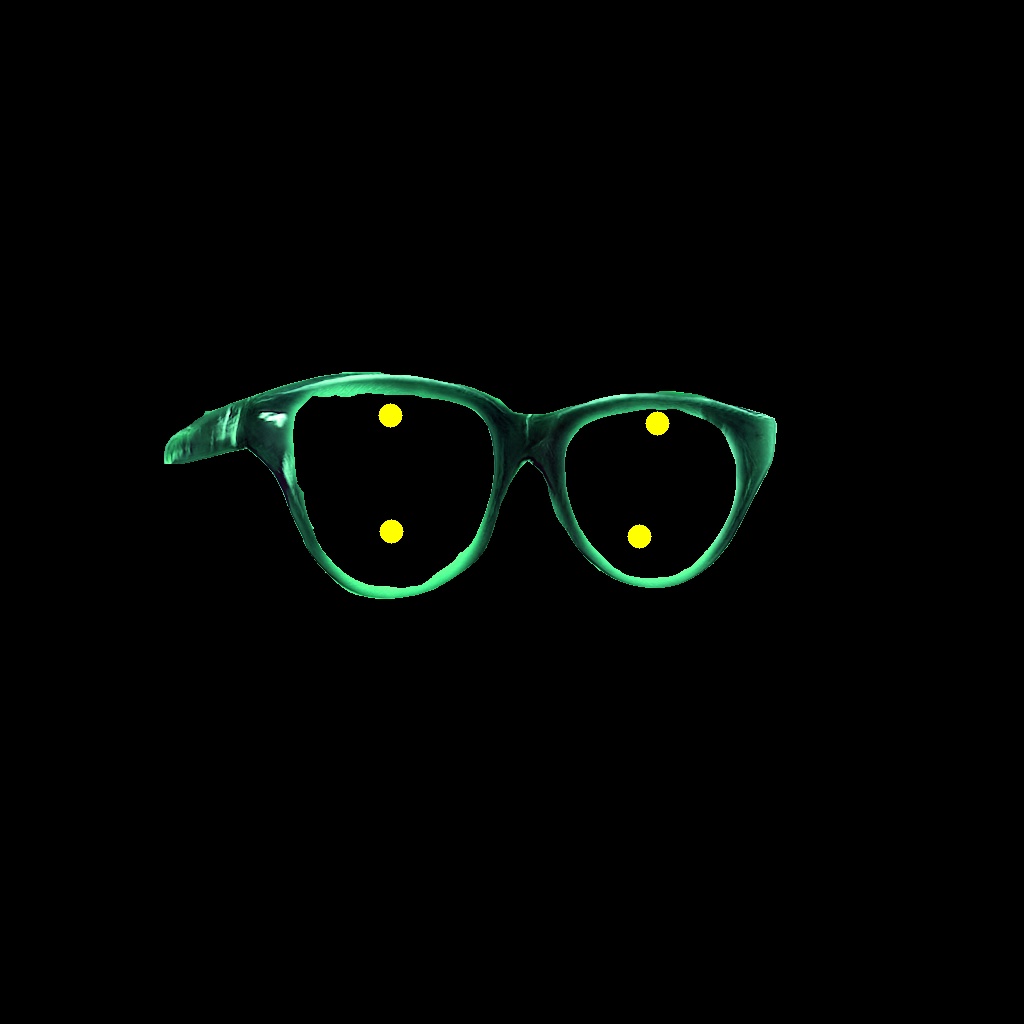}
     \end{subfigure}%
      \hspace{1mm}
     \begin{subfigure}[t]{0.3\columnwidth}
         \centering
         \includegraphics[width=2.5cm]{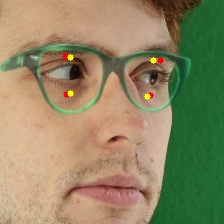}
     \end{subfigure}

     \caption{
     A rigid 2D transformation (incorporating rotation and translation) is applied to the glasses template by anchoring four facial landmarks. Landmark definitions are provided in \cref{fig:mediapipe}(b).
     }
     \label{fig:appendix_glasses}
\end{figure}

To synthesize pose-consistent glasses,
we fit a glasses template to the target face by minimizing the distance between two sets of anchor landmarks,
as illustrated in \cref{fig:appendix_glasses}.
The landmarks of each glasses template originate from the underlying face images in the head-pose-grid set (\cref{fig:glasses_aug} in the main paper).
Notably, the selected landmarks are positioned above and below the eye contours rather than directly on them. This ensures that the synthesis maintains independent of the subject's specific eye shape or gaze direction.
After the initial geometric fit, we further diversify the glasses appearance by randomized scale, frame color, opacity, and lens reflections.

\subsection{Mask Occlusion Synthesis} 

\begin{figure}[t]
     \centering
     \captionsetup[subfigure]{justification=centering, singlelinecheck=false}
     \scriptsize 
     \begin{subfigure}[t]{0.48\columnwidth}
         \centering
         \includegraphics[height=2.8cm]{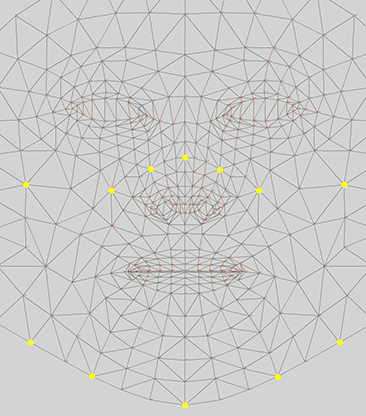}
         \caption{Face}
     \end{subfigure}
     \hfill
     \begin{subfigure}[t]{0.48\columnwidth}
         \centering
         \includegraphics[height=2.2cm]{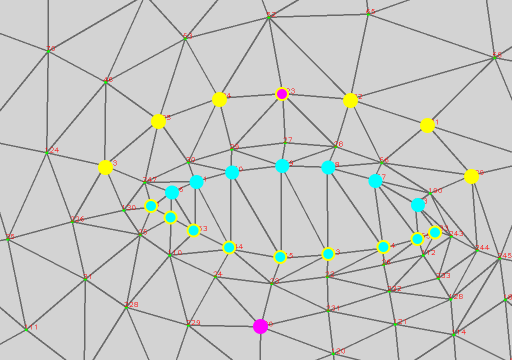}
         \caption{Left eye}
     \end{subfigure}

     \caption{
     (a) MediaPipe facial landmarks \cite{mediapipe};
     the region delineated by yellow points is used to simulate mask occlusion.
     (b) The ocular landmarks.
     Meganta points guide glasses synthesis (\cref{fig:appendix_glasses});
     yellow points define the eye region in the segmentation task, and cyan points specify the eye mask input $M$ for the iris-region generation algorithm (\cref{alg:segmap}).
     }
     \label{fig:mediapipe}
\end{figure}

To optimize computational efficiency during training, we generate facial mask regions for each training sample offline. As shown in \cref{fig:mediapipe}(a), the occlusion boundaries are initially defined by a polygon connecting 12 designated facial landmarks covering the nose and jawline. We further apply spline curve fitting to smoothen the polygon edges,
resulting in more realistic appearance and better preservation of facial geometry.

\begin{figure}[t]
     \centering
     \captionsetup[subfigure]{justification=centering, singlelinecheck=false}
     \scriptsize 
     
     \begin{subfigure}[t]{0.3\columnwidth}
         \centering
         \includegraphics[width=\textwidth]{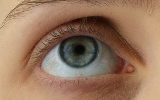}
         \caption{Looking up}
     \end{subfigure}%
     \hfill
     \begin{subfigure}[t]{0.3\columnwidth}
         \centering
         \includegraphics[width=\textwidth]{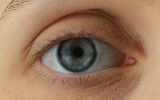}
         \caption{Looking straight}
     \end{subfigure}%
     \hfill
     \begin{subfigure}[t]{0.3\columnwidth}
         \centering
         \includegraphics[width=\textwidth]{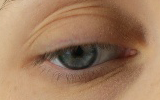}
         \caption{Looking down}
     \end{subfigure}

     \caption{
     The eyelid exhibits significant appearance variation correlated with the pitch component of gaze, $\phi$.
     }
     \label{fig:eye_sclera}
\end{figure}

\begin{figure}[t] 
     \centering
     \includegraphics[width=0.3\textwidth]{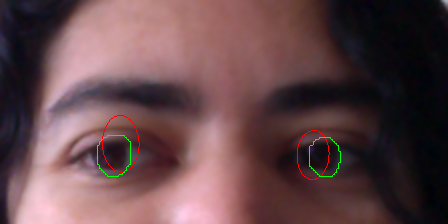}
     \caption{
     Segmentation masks generated by MediaPipe Iris (red) versus our proposed intensity-based method (green), highlighting the latter's superior alignment.
     }
     \label{fig:seg_compare}
\end{figure}
\begin{algorithm}[t]
\caption{Iris-region Mask Generation}
\label{alg:segmap}
    \begin{algorithmic}[1]
    \Require Eye image $I \in [0,1]^{h_I \times w_I \times 3}$, eye mask $M \in \{0,1\}^{h_I \times w_I}$, circular morphological kernel $K(\delta)$ of diameter $\delta$
    \State $Y \gets \text{Brightness}(I)$
    \State $Y \gets \text{GaussianBlur}(Y, \sigma=2)$ \Comment{Remove oculus reflections}
    \State $Y \gets Y + 0.5 \times \text{GaussianBlur}(1- \text{dilate} (M, K(\text{Mask\_width(} M \text{)}/6)), \sigma=15) $ 
    \State Clamp $Y$ to $[0,1]$ \Comment{Brighten the pixels near and outside the eye contour}
    \State $\tau \gets \text{Median}(\{ \text{pixels within } M \})$ \Comment{Brightness threshold}
    \State $M' \gets \text{Zeros\_like}(M)$
    \For{$M'_i$ in $M'$}
        \State $M'_i \gets Y_i < \tau$
    \EndFor \Comment{The raw iris mask}
    \State $M' \gets \text{MorphologicalOpen(}M', K(13)\text{)}$
    \State $M' \gets \text{MorphologicalClose(}M', K(5)\text{)}$ \Comment{Denoise by morphological operations}
    \State $M' \gets \text{ConnectedComponents}(M')[0]$ \Comment{Locate the largest connect component}
    \State $M' \gets \text{GaussianBlur}(M', \sigma=15) > 0.2 $ \Comment{Round the mask shape}
    \State \Return $M'$
    \end{algorithmic}
\end{algorithm}

\section{Automated Annotation Pipeline for Segmentation}

Although existing AGE works occasionally incorporate segmentation as an auxiliary task \cite{seg-ssl, pictorial}, they often rely on synthetic eye datasets like SynthesEyes \cite{syntheseyes}. These datasets suffer from a significant domain gap when applied to real-world, face-based AGE models.
To bridge this gap, we present an automated annotation pipeline capable of generating high-fidelity
eye and iris masks for arbitrary face datasets.
This pipeline serves as a general-purpose utility for face-related tasks beyond gaze estimation.

\begin{figure}[t]
    \centering
    \includegraphics[width=0.65\linewidth]{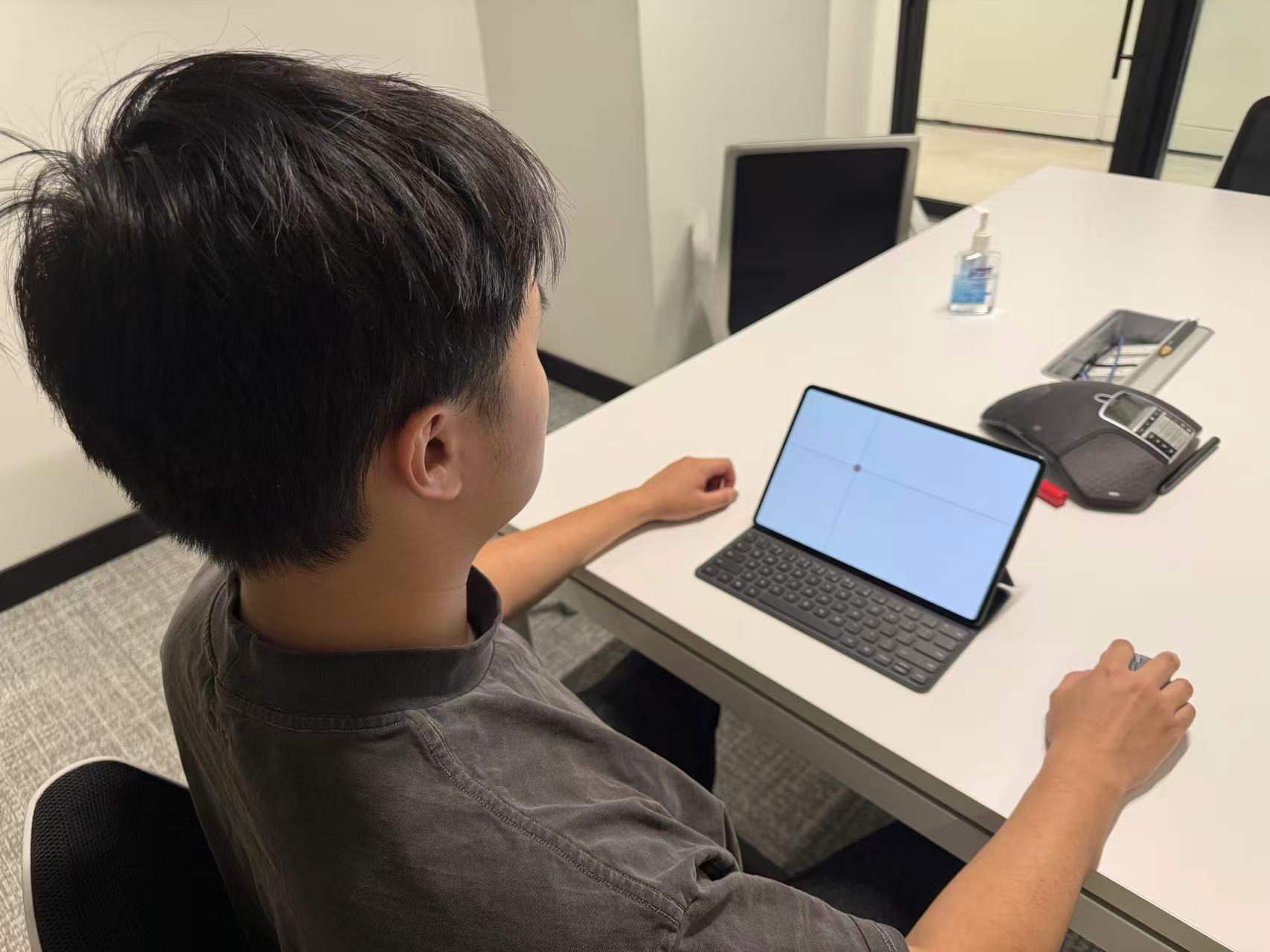}
    \caption{RealGaze data collection. A subject participating in session $a$ (no accessories, standard indoor lighting). } 
    \label{fig:realgaze_collection}
\end{figure}

\subsection{Eye Region} 
We utilize the MediaPipe \cite{mediapipe} face mesh that provides 478 fine-grained landmarks.
The eye-region mask is defined as a polygon encompassing the landmarks surrounding the eye and the immediate eyelid area
(see yellow landmarks in \cref{fig:mediapipe}(b)).
We specifically include the eyelid region because of its strong semantic correlation with the pitch component of gaze, $\phi$. As illustrated in \cref{fig:eye_sclera}, the projected eyelid area varies significantly with $\phi$; for instance, the eyelid occupies a larger visual area during downward gaze compared to upward gaze. By including this region in the segmentation task, 
we force the backbone to learn features that are sensitive to these subtle structural cues.

\begin{figure*}[t]
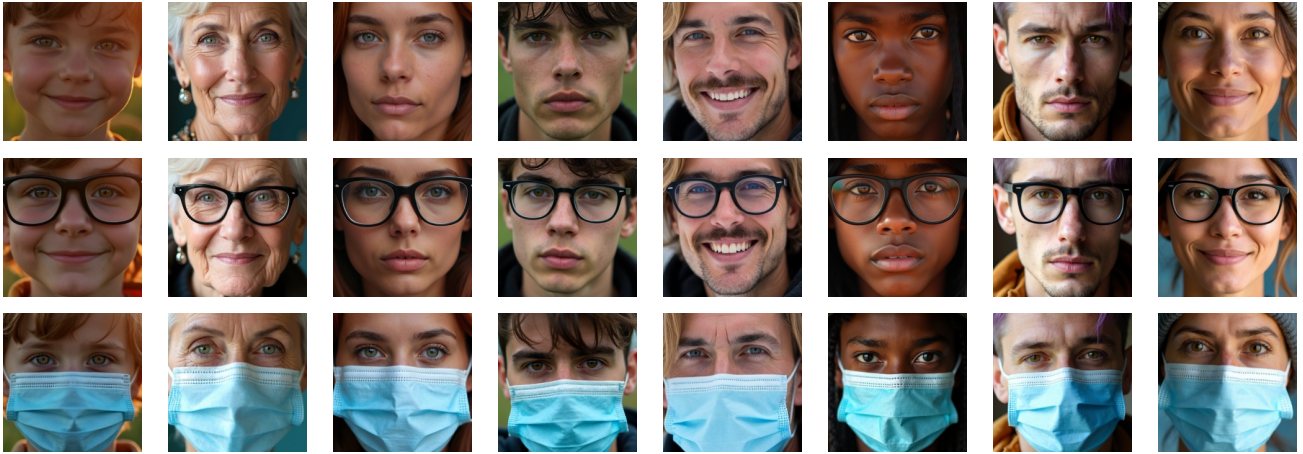

    \centering
    \foreach \img in {a,b,c,d,e,f,g,h} {
        \begin{minipage}{0.22\columnwidth}
            \centering
            \includegraphics[width=\columnwidth]{images/zerogaze_examples/vanilla/\img.png}
        \end{minipage}\hfill
    }
    
    \vspace{2mm} 
    
    \foreach \img in {a,b,c,d,e,f,g,h} {
        \begin{minipage}{0.22\columnwidth}
            \centering
            \includegraphics[width=\columnwidth]{images/zerogaze_examples/glasses/\img.png}
        \end{minipage}\hfill
    }

    \vspace{2mm}

    \foreach \img in {a,b,c,d,e,f,g,h} {
        \begin{minipage}{0.22\columnwidth}
            \centering
            \includegraphics[width=\columnwidth]{images/zerogaze_examples/masks/\img.png}
        \end{minipage}\hfill
    }
    \caption{Example ZeroGaze triplets($X, X_g, X_m$), generated via prompts $S^*$ (clean view), $S^* + S_1$ (glasses view), and $S^* + S_2$ (mask view). Note the high identity stability across diverse wearable prompts.
    }
    \label{fig:appendix_triplets}
\end{figure*}
\begin{figure*}[t]
    \centering
    \captionsetup[subfigure]{justification=centering, singlelinecheck=false}

    \begin{minipage}[c]{0.2\linewidth}
        \centering
        \textbf{Excessive\\ Appearance\\ Alteration}
    \end{minipage}\hspace{1mm}
    \begin{subfigure}[c]{0.18\linewidth}
        \centering
        \includegraphics[width=2.6cm]{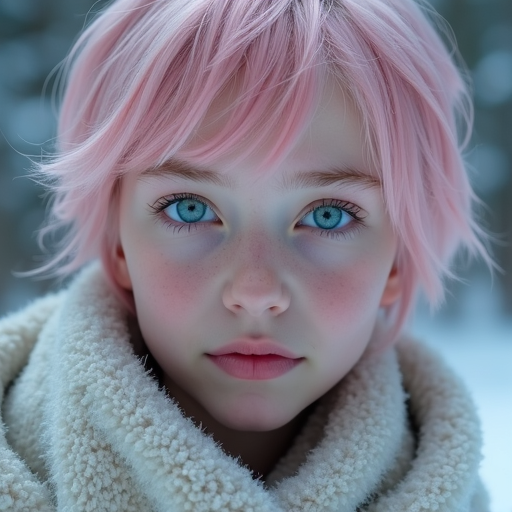}
        \caption*{Original}
    \end{subfigure}\hspace{1mm}
    \begin{subfigure}[c]{0.18\linewidth}
        \centering
        \includegraphics[width=2.6cm]{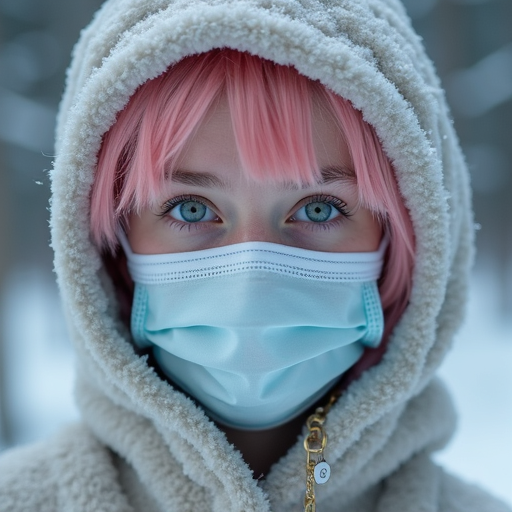}
        \caption*{Masked}
    \end{subfigure}\hspace{1mm}
    \begin{subfigure}[c]{0.18\linewidth}
        \centering
        \includegraphics[width=2.6cm]{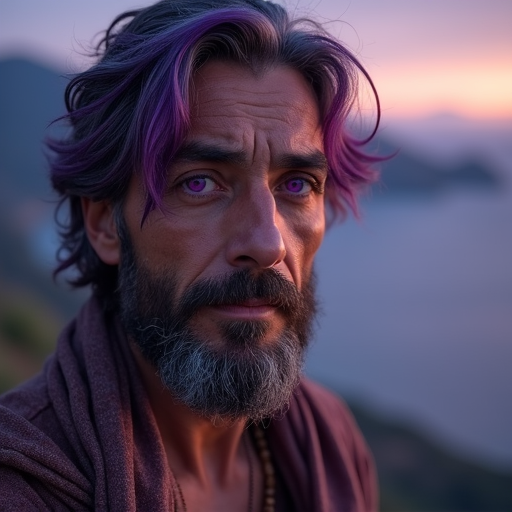}
        \caption*{Original}
    \end{subfigure}\hspace{1mm}
    \begin{subfigure}[c]{0.18\linewidth}
        \centering
        \includegraphics[width=2.6cm]{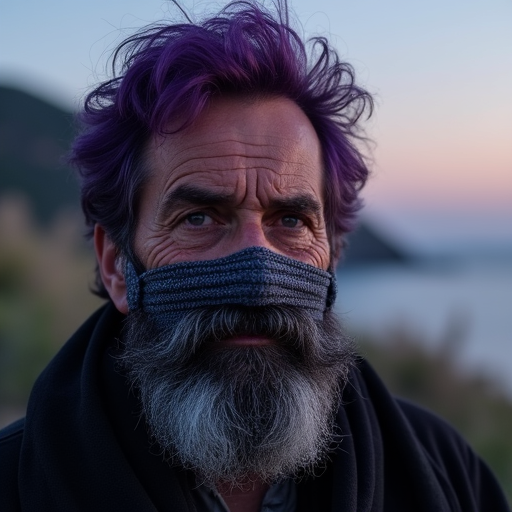}
        \caption*{Masked}
    \end{subfigure}

    \vspace{1mm} 

    \begin{minipage}[c]{0.2\linewidth}
        \centering
        \textbf{Inconsistent\\ Identity}
    \end{minipage}\hspace{1mm}
    \begin{subfigure}[c]{0.18\linewidth}
        \centering
        \includegraphics[width=2.6cm]{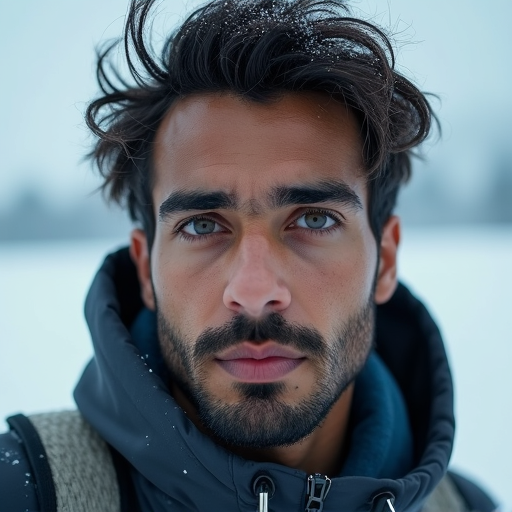}
        \caption*{Original}
    \end{subfigure}\hspace{1mm}
    \begin{subfigure}[c]{0.18\linewidth}
        \centering
        \includegraphics[width=2.6cm]{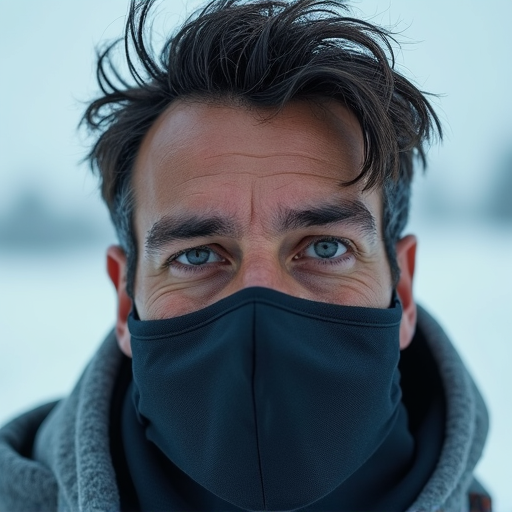}
        \caption*{Masked}
    \end{subfigure}\hspace{1mm}
    \begin{subfigure}[c]{0.18\linewidth}
        \centering
        \includegraphics[width=2.6cm]{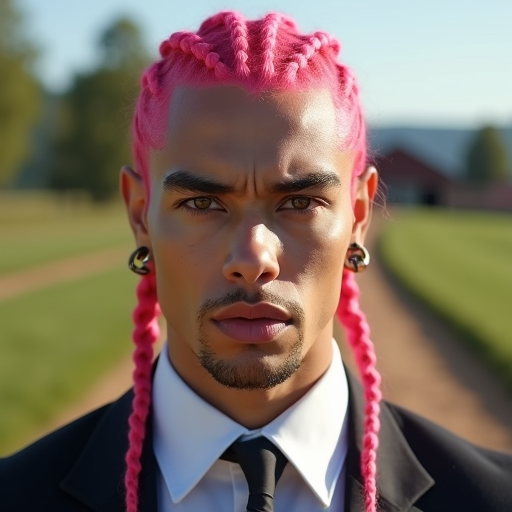}
        \caption*{Original}
    \end{subfigure}\hspace{1mm}
    \begin{subfigure}[c]{0.18\linewidth}
        \centering
        \includegraphics[width=2.6cm]{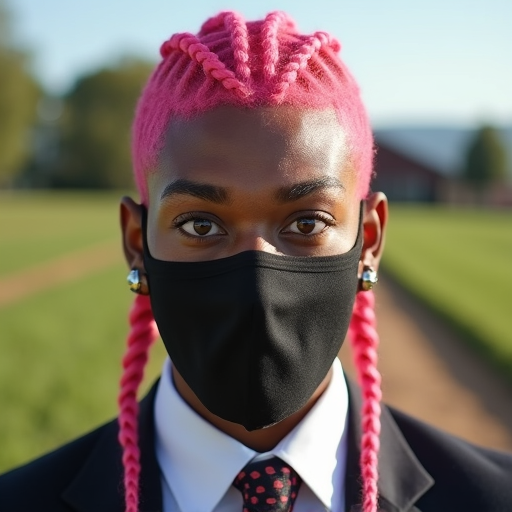}
        \caption*{Masked}
    \end{subfigure}

    \caption{Omitting the ``\texttt{medical}'' keyword in $S_2$ induces 
    undesired artifacts (\eg, apparel change or non-realistic occlusion) and catastrophic identity shifts (\eg, changes in age and gender).
    Images are shown at raw $512 \times 512$ resolution prior to cropping.}  
    \label{fig:zerogaze_mask}
\end{figure*}
\begin{figure}[h]
    \centering
    \begin{subfigure}[c]{0.25\linewidth} 
        \centering
        \includegraphics[width=\textwidth]{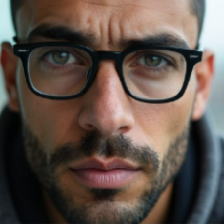}
        \caption*{Original}
    \end{subfigure}
    \hspace{2mm}
    \begin{subfigure}[c]{0.25\linewidth}
        \centering
        \includegraphics[width=\textwidth]{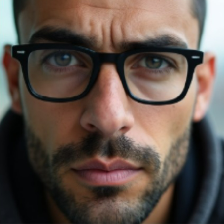}
        \caption*{Glasses}
    \end{subfigure}

    \vspace{2mm} 

    \begin{subfigure}[c]{0.25\linewidth}
        \centering
        \includegraphics[width=\textwidth]{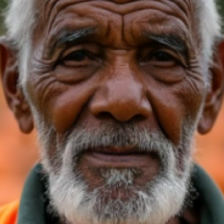}
        \caption*{Original}
    \end{subfigure}
    \hspace{2mm}
    \begin{subfigure}[c]{0.25\linewidth}
        \centering
        \includegraphics[width=\textwidth]{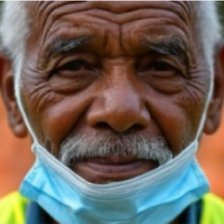}
        \caption*{Masked}
    \end{subfigure}

    \caption{Samples of ($X, X_g$), ($X, X_m$) pairs with top $5\%$ similarity scores. Excessively high similarity scores may imply improper augmentation.}  
    \label{fig:zerogaze10K}
\end{figure} 
\begin{figure*}[tb]
  \centering
  \begin{minipage}{0.02\linewidth}
    \rotatebox{90}{\small ~~~ETH-XGaze}
  \end{minipage}
  \begin{minipage}{0.48\linewidth}
    \centering
    \includegraphics[width=\linewidth]{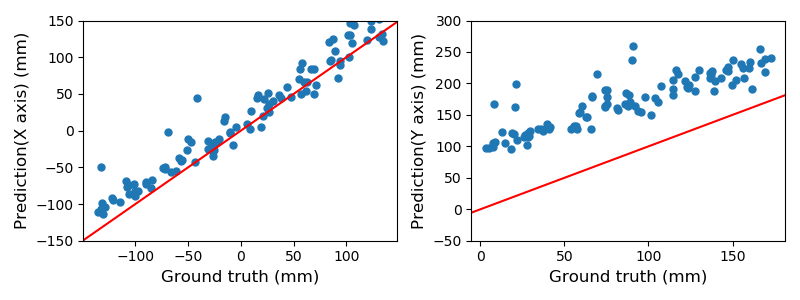}\\
  \end{minipage}
  \hfill
  \begin{minipage}{0.48\linewidth}
    \centering
    \includegraphics[width=\linewidth]{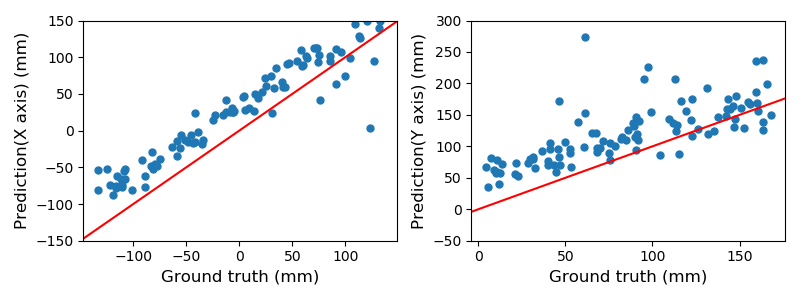}\\
  \end{minipage}
  
  
  \begin{minipage}{0.02\linewidth}
    \rotatebox{90}{\small ~~~~~~Ours(MobileNet)} 
  \end{minipage}
  \begin{minipage}{0.48\linewidth}
    \centering
    \includegraphics[width=\linewidth]{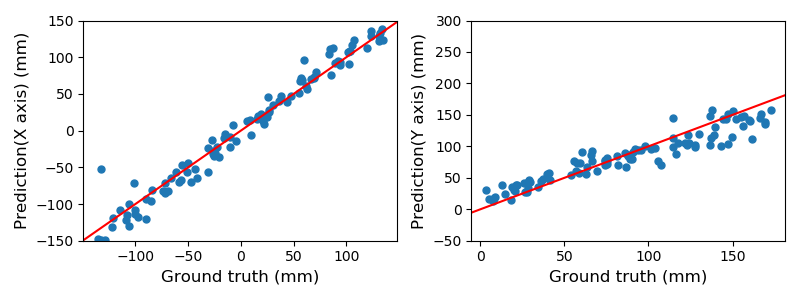}\\
    \subcaption{Session $b$ \\ (no accessories, standard lighting)}
  \end{minipage}
  \hfill
  \begin{minipage}{0.48\linewidth}
    \centering
    \includegraphics[width=\linewidth]{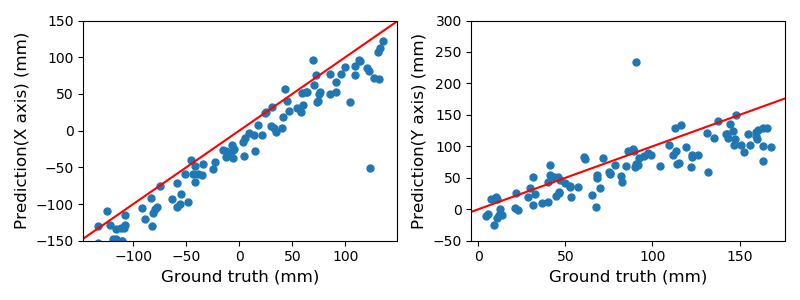}\\
    \subcaption{Session $c$ \\ (glasses, standard lighting)}
  \end{minipage}
  \caption{
  Scatter plots of ground-truth labels (X-axis) versus model predictions (Y-axis) for a representative RealGaze subject. The persistent linear trend validates our training-free calibration approach. The red line denotes the identity mapping $y=x$.
  }
  \label{fig:scatter}
\end{figure*}

\subsection{Iris Region}
While MediaPipe Iris is available to provide five landmarks per iris,
we observe that its output is insufficiently precise for per-pixel segmentation. As shown in \cref{fig:seg_compare}, MediaPipe's estimates are frequently oversized and centered inaccurately relative to the true iris.
To address this,
we propose an image-processing-based alternative (\cref{alg:segmap})
grounded in the physical prior that the iris is significantly darker than the surrounding sclera.
Given an inner-eye-region mask (eyelid excluded, defined by the cyan landmarks in \cref{fig:mediapipe}(b)), we identify the darkest area therein by thresholding the brightness value, and 
apply morphological operations to yield a contiguous, rounded iris shape.
To prevent supervision from erroneous annotation, we calculate the Intersection over Union (IoU) between the predicted eye and iris masks. Samples with an $IoU < 0.2$ are considered unreliable and discarded.
During training, the segmentation loss $L_{seg}$ is dynamically disabled for samples without segmentation labels.

\section{Benchmark Dataset Details}

\subsection{RealGaze} 
In RealGaze data collection, participants utilize a 13-inch tablet in a seated position, maintaining their natural head-movement habits (\cref{fig:realgaze_collection}).
To capture precise gaze coordinates, we develop a dedicated application that renders a randomized red target dot. Subjects are instructed to gaze at the target and simultaneously click it with a mouse. Each click triggers a $1920 \times 1080$ frame capture, pairing the visual data with the target's screen coordinates. A brief temporal delay is implemented between target shifts to prevent motion blur.
We also perform face and landmark detection on the capture, so that those exhibiting eye-blinks or significant motion blur are automatically discarded.

Of the 20 participants, 6 require eyeglasses to resolve the screen targets. As a result, they participate in six specific sessions (instead of the nine sessions for others), spanning combinations of wearables (glasses, glasses and mask) and three illumination conditions (standard, low-light, and harsh side-lighting). These sessions are counted in the Overall setting in the RealGaze experiments (\cref{s:exp_realgaze}), but not included in the remaining 4 settings (Ideal, Side-lit, Glasses, Masks) for fair comparison with Ideal.

\subsection{ZeroGaze} 

We use Flux.1-Dev \cite{flux} to generate high-resolution face images from $140,000$
base prompts sourced from the SFHQ dataset \cite{sfhq}.
To encourage consistent zero head pose and gaze, we normalize each base prompt by adding the alignment prompt
$S^* =$ ``\texttt{full frontal portrait with strict eye contact with the camera lens, eyes centered and head forward (0°), tight face shot}'' 
and removing conflicting descriptors related to gaze, head pose, or accessories.
Keywords not targeting photorealistic face generation (\eg, non-realistic artistic styles) are also trimmed.

For each normalized prompt, we generate a clean view $X$ and two augmented views, $X_g$ and $X_m$,
by appending the prompts $S_1 =$ ``\texttt{wearing a pair of black glasses, position glasses so each eye is in the middle of each lens}'' and $S_2 =$ ``\texttt{wearing a medical mask covering the bottom of the face}'', respectively.
The resulting triplets are designed to differ only in the presence of wearables, as shown in \cref{fig:appendix_triplets}.
We empirically find that the keyword ``\texttt{medical}'' in $S_2$ is essential for preventing unnatural facial artifacts and unintended identity changes (\cref{fig:zerogaze_mask}).

To assess identity consistency within each triplet,
we extract normalized facial identity features using the pre-trained ArcFace model \cite{arcface}, denoted by $F_{\mathrm{ID}}(\cdot)$, and compute the pairwise cosine similarities: 
$S(X, X_k) := \cos(F_{\mathrm{ID}}(X), F_{\mathrm{ID}}(X_k)), k \in \{g, m\}$. 
Low similarity scores indicate identity alteration, whereas excessively high scores often indicate ineffective augmentation (see \cref{fig:zerogaze10K}).
We therefore discard triplets whose $S(X, X_g)$ or $S(X, X_m)$ falls within the highest 5\% of similarity scores.
Among the remaining candidates,
we select 10,000 triplets with the highest average similarity across the two pairs. 
Together, the head pose and identity filters help isolate wearable-induced appearance changes, enabling ZeroGaze to assess a model's ability to marginalize gaze-irrelevant features.

\begin{table*}[tb]
  \caption{Results of personalized calibration on RealGaze using different numbers of calibration points (errors in mm).
  }
  \label{tab:calib_alex2}
  \centering
  
  \resizebox{0.9\textwidth}{!}{%
  \begin{threeparttable}
      \begin{tabular}{@{}l | ccc | rrr | rrr | rrr | rrr@{}}
        \toprule
        \textbf{Model} & \multicolumn{3}{c}{\textbf{Overall}} & \multicolumn{3}{c}{\textbf{Ideal}} & \multicolumn{3}{c}{\textbf{Side-Lit}} & \multicolumn{3}{c}{\textbf{Glasses}} & \multicolumn{3}{c}{\textbf{Masks}} \\
     Calibration Setting & \multicolumn{1}{c}{$d_X$} & \multicolumn{1}{c}{$d_Y$} & \multicolumn{1}{c}{$\|d\|_2$} & \multicolumn{1}{c}{$d_X$} & \multicolumn{1}{c}{$d_Y$} & \multicolumn{1}{c}{$\|d\|_2$} & \multicolumn{1}{c}{$d_X$} & \multicolumn{1}{c}{$d_Y$} & \multicolumn{1}{c}{$\|d\|_2$} & \multicolumn{1}{c}{$d_X$} & \multicolumn{1}{c}{$d_Y$} & \multicolumn{1}{c}{$\|d\|_2$} & \multicolumn{1}{c}{$d_X$} & \multicolumn{1}{c}{$d_Y$} & \multicolumn{1}{c}{$\|d\|_2$} \\
        \midrule \midrule
        \multicolumn{16}{l}{\textbf{ETH-XGaze (ResNet-50, 25.6M)}} \\
        Uncalibrated & 42.9 & 59.7 & 82.1 & 24.8 & 62.7 & 72.3 & 33.0 & 56.9 & 71.8 & 73.3 & 56.6 & 104.8 & 31.3 & 62.0 & 75.1\\
        1 Point & 29.0 & 28.0 & 45.5 & 14.9 & 21.1 & 28.6 & 14.7 & 19.7 & 27.2 & 42.4 & 30.2 & 58.1 & 25.5 & 33.6 & 47.2\\
        5 Points & 23.3 & 22.6 & 36.3 & 14.3 & 18.1 & 25.7 & 15.1 & 17.7 & 25.8 & 33.7 & 25.2 & 46.9 & 24.7 & 24.6 & 38.9\\
        \midrule
        \multicolumn{16}{l}{UniGaze-B-Joint (ViT-B, 86.6M)} \\
        Uncalibrated & 21.1 & 44.4 & 52.8 & 16.1 & 33.7 & 40.6 & 17.4 & 35.0 & 41.8 & 26.9 & 52.9 & 63.8 & 19.7 & 40.2 & 48.6\\
        1 point & 16.6 & 20.9 & 29.8 & 12.6 & 18.7 & 25.1 & 13.0 & 18.6 & 25.3 & 21.4 & 22.9 & 35.2 & 14.7 & 21.9 & 29.2\\
        5 points & 14.3 & 18.3 & 26.0 & 10.9 & 15.9 & 21.2 & 12.4 & 15.3 & 22.0 & 19.4 & 19.7 & 30.9 & 14.0 & 18.9 & 26.2\\
        \midrule
        \multicolumn{16}{l}{Ours (MobileNet, 3.8M)} \\
        Uncalibrated & 22.3 & 34.9 & 46.3 & 16.0 & 28.8 & 36.6 & 18.6 & 28.5 & 37.0 & 25.4 & 33.4 & 46.6 & 20.9 & 36.0 & 45.3\\
        1 Point & 19.9 & 23.8 & 33.5 & 15.2 & 20.5 & 28.4 & 14.2 & 21.0 & 28.1 & 23.7 & 24.0 & 37.5 & 19.9 & 27.0 & 37.8\\
        5 Points & 17.3 & 21.5 & 29.7 & 12.5 & 17.6 & 24.0 & 13.6 & 16.9 & 24.3 & 21.0 & 21.4 & 33.5 & 16.2 & 25.2 & 33.1\\
        \midrule
        \multicolumn{16}{l}{Our (ViT-B, 86.6M)} \\
        Uncalibrated & 19.1 & 35.8 & 44.4 & 14.9 & 29.3 & 36.2 & 17.7 & 27.9 & 35.9 & 24.4 & 33.5 & 46.0 & 18.8 & 36.5 & 45.3\\
        1 Point & 16.8 & 17.8 & 27.4 & 11.7 & 14.0 & 20.3 & 12.7 & 15.1 & 22.0 & 23.9 & 19.4 & 34.1 & 13.3 & 19.1 & 25.9\\
        5 Points & 14.7 & 15.9 & 24.2 & 10.4 & 12.2 & 17.8 & 11.6 & 13.5 & 19.8 & 20.4 & 17.9 & 30.2 & 13.9 & 16.6 & 24.3 \\
      \bottomrule
      \end{tabular}
      \end{threeparttable}
      }
\end{table*}
\begin{table*}[tb]
  \caption{
  Results of personalized calibration on MPIIFaceGaze $D_M$ using different numbers of calibration points (errors in mm).
  }
  \label{tab:gaze322_comparison}
  \centering
  \resizebox{\textwidth}{!}{%
    \begin{threeparttable}
      \begin{tabular}{@{}l c | rrr | rrr | rrr | rrr | rrr@{}}
        \toprule
        \textbf{Model} & \multicolumn{1}{c}{\textbf{Model}} & \multicolumn{3}{c}{\textbf{Uncalibrated}} & \multicolumn{3}{c}{\textbf{3 Pts}} & \multicolumn{3}{c}{\textbf{5 Pts}} & \multicolumn{3}{c}{\textbf{10 Pts}} & \multicolumn{3}{c}{\textbf{20 Pts}} \\
     Error (mm) & \multicolumn{1}{c}{\textbf{size}} & \multicolumn{1}{c}{$d_X$} & \multicolumn{1}{c}{$d_Y$} & \multicolumn{1}{c}{$\|d\|_2$} & \multicolumn{1}{c}{$d_X$} & \multicolumn{1}{c}{$d_Y$} & \multicolumn{1}{c}{$\|d\|_2$} & \multicolumn{1}{c}{$d_X$} & \multicolumn{1}{c}{$d_Y$} & \multicolumn{1}{c}{$\|d\|_2$} & \multicolumn{1}{c}{$d_X$} & \multicolumn{1}{c}{$d_Y$} & \multicolumn{1}{c}{$\|d\|_2$} & \multicolumn{1}{c}{$d_X$} & \multicolumn{1}{c}{$d_Y$} & \multicolumn{1}{c}{$\|d\|_2$} \\
        \midrule\midrule

        Gaze322 (ResNet-18) & 11.4M & - & - & 101.9 & - & - & 79.3$^*$  & - & - & 62.3$^*$ & - & - & 56.7 & - & - & 45.4$^*$ \\
        ETH-XGaze (ResNet-50) & 25.6M & 47.4 & 82.9 & 104.3 & 32.1 & 63.0 & 76.6 & 26.6 & 45.6 & 57.0 & 23.3 & 47.8 & 56.5 & 21.3 & 46.3 & 54.8 \\
        UniGaze-B-Joint & 86.6M & 36.6 & 77.7 & 92.4 & 27.6 & 54.3 & 66.5 & 18.6 & 53.3 & 60.3 & 18.5 & 47.5 & 54.3 & 18.8 & 45.7 & 52.8\\
        Ours (MobileNet) & 3.8M & 39.8 & 70.8 & 88.1 & 29.8 & 51.6 & 63.6 & 25.6 & 50.9 & 59.3 & 21.1 & 47.4 & 54.8 & 19.6 & 45.6 & 52.6 \\
        Ours (ViT-B) & 86.6M & 40.4 & 73.2 & 90.7 & 27.2 & 56.6 & 67.6 & 18.9 & 49.4 & 56.2 & 19.2 & 47.9 & 54.7 & 20.8 & 46.0 & 53.2 \\
    
      \bottomrule
      \end{tabular}
    \begin{tablenotes}
    \footnotesize  
    \item[*] Manually measured based on Fig. 5 in \cite{gaze322}.
    \end{tablenotes}
  \end{threeparttable}
  }
\end{table*}

\section{Training-free Personalized Calibration}

Although our framework achieves state-of-the-art generalization, the intrinsic physiological variance of the human eye (\ie, angle kappa), imposes a ``glass ceiling'' on zero-shot precision. Recall our RealGaze experiments (\cref{s:exp_realgaze}): on a standard 13-inch tablet, generalization errors can still exceed 40 mm, which is sufficient for coarse interaction but inadequate for fine-grained tasks.
To bridge this gap, we propose a lightweight, training-free calibration mechanism. Unlike recent optimization-based methods like Gaze322 \cite{gaze322}, which require over 10 points (thus a long, disturbing calibration process per use) and significant test-time computation, our approach leverages the linear relationship between model estimations and 2D ground truths. As illustrated in \cref{fig:scatter}, this linear pattern remains remarkably consistent on both X and Y axes unless the model is catastrophically occluded.
We model the output-label relationship as a 2D linear system with person-specific slopes and intercepts.
For 1-point calibration, we estimate intercepts only; given more calibration points, we perform linear regression on them to determine both slopes and intercepts.

\subsection{Experiments on RealGaze}
\begin{table*}[t]
  \caption{
  Cross-domain evaluation results on MPIIFaceGaze $D_M$ and EyeDiap $D_E$ (angular errors in degrees; $d_\phi$ and $d_\psi$ are the pitch and yaw components respectively).
  }
  \label{tab:oldschool}
  \centering
  \resizebox{0.8\textwidth}{!}{%
  \begin{threeparttable}
  \begin{tabular}{@{}lc|ccc|ccc@{}}

    \toprule

    Model (Backbone) & Model & \multicolumn{3}{c}{$\rightarrow D_M$} & \multicolumn{3}{c}{$\rightarrow D_E$} \\
    & Size & \multicolumn{1}{c}{$d$} & \multicolumn{1}{c}{$d_\phi$} & \multicolumn{1}{c}{$d_\psi$} & \multicolumn{1}{c}{$d$} & \multicolumn{1}{c}{$d_\phi$} & \multicolumn{1}{c}{$d_\psi$} \\
    \midrule

    (a) Trained on $D_X$ only & & & \\


    ETH-XGaze (ResNet-50) \cite{xgaze} & 25.6M & 6.94 (7.50$^*$) & 5.15 & 3.75 & 8.87 (11.0$^*$) & 5.23 & 6.40\\

    PureGaze (ResNet-50) \cite{puregaze} & 31M & 6.85 (7.08$^*$) & 4.91 & 3.89 & 7.40 (7.48$^*$) & 4.67 & 4.75 \\


    CDG (ResNet-50) \cite{crga} & 25.6M & 6.73$^*$ &-&-& 7.95$^*$ &-&- \\




    FSCI (ResNet-18) \cite{deconfounded} & 16.4M & 6.93 (5.79$^*$) & 4.64 & 4.32 & 7.94 (6.96$^*$) & 6.12 & 3.88 \\

    UniGaze-B-16 (ViT-B) \cite{unigaze} & 86.6M & 6.09 (6.21$^*$) & 5.02 & 2.80 & 6.63 (6.64$^*$) & 4.86 & 3.67 \\

    UniGaze-H-14-\texttt{CrossX} (ViT-H) & 632M & 5.82 (5.57$^*$) & 4.75 & 2.52 & 5.80 (6.53$^*$) & 4.26 & 3.09 \\

    Ours (MobileNet) & 3.8M & 6.04 & 4.25 & 3.53 & 7.64 & 6.04 & 3.68 \\ 

    Ours (ViT-B) & 86.6M & 5.25 & 3.82 & 2.78 & 6.76 & 4.92 & 3.51 \\ 

    \midrule

    (b) Trained on multiple datasets   \\




    UniGaze-B-16-Joint$^\dagger$ (\text{@} 5 datasets) & 86.6M & 5.35 & 3.95 & 2.78 & 5.02 (5.52$^*$) & 3.67 & 2.69\\


    
    UniGaze-H-14-Joint$^\dagger$ (\text{@} 5 datasets) & 632M & 5.08 & 3.86 & 2.54 & 4.53 (5.16$^*$) & 3.25 & 2.45\\

    Ours (MobileNet $@\{D_X$,$D_C$,$D_N\}$) & 3.8M & 5.16 & 3.84 & 2.75 & 6.47 & 4.86 & 3.21 \\ 

    Ours (ViT-B $@\{D_X$,$D_C$,$D_N\}$) & 86.6M & 4.57 & 3.33 & 2.50 & 5.51 & 3.88 & 3.09 \\
    \midrule
  \bottomrule
  \end{tabular}

    \begin{tablenotes}

    \footnotesize  

    \item[$\dagger$] The models are evaluated on a portion of dataset, because the rest is used in training; these models benefit from a smaller training-test domain gap. $^*$ Results reported in the original paper.

    \end{tablenotes}
\end{threeparttable}
}
\end{table*}
\begin{table*}[t]
  \caption{Ablation studies on RealGaze (errors in mm).
  }
  \label{tab:appendix_ablation}
  \centering
  \resizebox{\textwidth}{!}{%
      \begin{tabular}{@{}l rrr | rrr | rrr | rrr | rrr@{}}
        \toprule
        \textbf{Model} & \multicolumn{3}{c}{\textbf{Overall}} & \multicolumn{3}{c}{\textbf{Ideal}} & \multicolumn{3}{c}{\textbf{Side-Lit}} & \multicolumn{3}{c}{\textbf{Glasses}} & \multicolumn{3}{c}{\textbf{Masks}} \\
         & \multicolumn{1}{c}{$d_X$} & \multicolumn{1}{c}{$d_Y$} & \multicolumn{1}{c}{$\|d\|_2$} & \multicolumn{1}{c}{$d_X$} & \multicolumn{1}{c}{$d_Y$} & \multicolumn{1}{c}{$\|d\|_2$} & \multicolumn{1}{c}{$d_X$} & \multicolumn{1}{c}{$d_Y$} & \multicolumn{1}{c}{$\|d\|_2$} & \multicolumn{1}{c}{$d_X$} & \multicolumn{1}{c}{$d_Y$} & \multicolumn{1}{c}{$\|d\|_2$} & \multicolumn{1}{c}{$d_X$} & \multicolumn{1}{c}{$d_Y$} & \multicolumn{1}{c}{$\|d\|_2$} \\
        \midrule\midrule
        (a) Variance Loss \\
        Ours (MobileNet, baseline) & 22.3 & 34.9 & 46.3 & 16.0 & 28.8 & 36.6 & 18.6 & 28.5 & 37.0 & 25.4 & 33.4 & 46.6 & 20.9 & 36.0 & 45.3\\

        Replacing $\tau=0.5$ with variance loss & 21.9 & 38.0 & 47.9 & 17.0 & 33.9 & 41.2 & 22.2 & 28.1 & 39.6 & 24.9 & 42.6 & 54.3 & 18.8 & 40.7 & 48.6
\\ \midrule
        (b) Pre-training \\

        Ours (ViT-B, MAE-pretrained) & 19.1& 35.8& 44.4& 14.9& 29.3& 36.2& 17.7& 27.9& 35.9& 24.4& 33.5& 46.0 & 18.8& 36.5& 45.3
\\

        ViT-B, UniGaze-pretrained & 18.8 & 36.3 & 44.6 & 14.3 & 32.2 & 37.9 & 18.9 & 29.2 & 37.8 & 23.3 & 37.1 & 47.9 & 17.0 & 41.8 & 49.2

\\ \midrule

        (c) Data Filtering \\

        Ours (MobileNet, baseline) & 22.3 & 34.9 & 46.3 & 16.0 & 28.8 & 36.6 & 18.6 & 28.5 & 37.0 & 25.4 & 33.4 & 46.6 & 20.9 & 36.0 & 45.3\\

        Ours (MobileNet, $\simeq 0$ filtering) & 24.9 & 39.7 & 51.7 & 17.7& 26.7 & 35.3 & 19.6 & 34.0 & 43.3 & 32.5 & 38.1 & 55.8 & 24.7 & 44.3 & 55.7

\\ \hdashline 

        UniGaze-B-Joint (ViT-B) \cite{unigaze} & 21.1 & 44.4 & 52.8 & 16.1 & 33.7 & 40.6 & 17.4 & 35.0 & 41.8 & 26.9 & 52.9 & 63.8 & 19.7 & 40.2 & 48.6
\\

        Ours (ViT-B, baseline) & 19.1& 35.8& 44.4& 14.9& 29.3& 36.2& 17.7& 27.9& 35.9& 24.4& 33.5& 46.0 & 18.8& 36.5& 45.3
\\
        Ours (ViT-B, $\simeq 0$ filtering) & 21.4 & 35.0 & 46.1 & 15.5 & 32.2 & 39.8 & 17.0 & 29.1 & 38.7 & 28.7 & 35.3 & 50.9 & 18.7 & 35.2 & 43.7\\
        
      \bottomrule
      \end{tabular}
    }
\end{table*}

We test the proposed calibration method on each session of RealGaze, under two settings:
\begin{itemize}
    \item \textbf{1-Point Calibration}: We determine a single calibration point by averaging the labels and model predictions of the three samples whose ground truth PoG are closest to the screen center. 
    We then report the average 2D gaze errors using the intercept-calibrated results of the remaining samples.
    In practical deployment, the real-time inference speed of our lightweight model makes it possible to combine multiple temporally adjacent predictions to stabilize calibration point collection, with negligible delay.
    \item \textbf{5-Point Calibration}: By repeating the process above, we collect five calibration points (screen center, and the four screen corners), and report the errors on the remaining samples calibrated by slopes and intercepts.
\end{itemize}
Results in \cref{tab:calib_alex2} demonstrate that a single calibration point provides the most significant performance leap, particularly in mitigating the pitch bias identified in our ZeroGaze experiments (\cref{s:exp_zerogaze}). While 5-point calibration offers further gains, the marginal utility decreases, suggesting that 1-point calibration is the sweet spot for balancing accuracy and user experience.
Notably, in cases where the model is severely challenged (\eg, the Glasses sessions), calibration is less effective at correcting the errors (especially for ETH-XGaze), indicating that while calibration fixes bias, it cannot fully compensate for a loss of feature signal.

\subsection{Experiments on $D_M$}

We also perform calibration testing on $D_M$ to provide a head-to-head comparison with Gaze322 \cite{gaze322}, following their evaluation protocol.
Given a calibration point count $n_C$,
we randomly draw $n_C$ samples for each subject (without using closest neighbors in the earlier RealGaze experiments, for fairness). To account for the randomness of the point selection, we repeat the process $9$ times and report the median 2D gaze errors.
As $D_M$ does not have official 3D-to-PoG pipeline, we map the output 3D gaze to PoG using the annotated HPE labels and camera parameters, which may introduce errors, resulting a higher baseline 2D error across all methods compared to RealGaze, although the 3D angular errors appear to be low (see \cref{s:appendix_old}).

The results in \cref{tab:gaze322_comparison} show that our training-free calibration method achieves a similar or marginally larger performance gain when the calibration point count is low ($n_C \leq 10$). Notably, our method does not require exclusive test-time training, memory overhead, or the computational cost of optimization. Gaze322 only demonstrates a clear advantage when the calibration point count significantly exceeds 10, a threshold that is often impractical for real-world user experience.

\section{More Experimental Results}
\subsection{Conventional Cross-Dataset Experiment}
\label{s:appendix_old}

To situate our work within the existing literature, we evaluate our models on two widely used unseen datasets: MPIIFaceGaze $D_M$ \cite{mpiiface} and EYEDIAP $D_E$ \cite{eyediap}.
We consider two training settings: training on $D_X$ alone and joint training on multiple datasets.
Since $D_E$ lacks official NCCS 3D labels, we adopt the labels provided by UniGaze \cite{unigaze}.

To improve comparability with prior methods, we omit the data filtering described in \cref{s:setup} and the classification loss. We retain data resampling, eye-region segmentation, augmentation, and the SupCon objectives, except that $L_{\phi}^S$ is omitted when training on $D_X$ alone.
\cref{tab:oldschool} reports mean angular errors alongside results from recent domain generalization methods, including
CDG \cite{crga}, FSCI \cite{deconfounded}, and UniGaze \cite{unigaze}.
For methods with publicly available checkpoints or sufficiently detailed implementations, we reproduce their evaluations on the target datasets and report the pitch and yaw error components, together with the overall angular error. We also include the results reported in the original papers.
Some reproduced results differ noticeably from the reported values, particularly on $D_E$, for which gaze label generation varies across studies (as discussed in \cref{s:app_datasets}).
Our re-evaluations provide a comparison under a common evaluation protocol.
However, the UniGaze-Joint results require separate consideration: as noted in \cref{tab:oldschool}, these models are evaluated on held-out portions of datasets that also contribute to their training and therefore benefit from a smaller training–test domain gap.


Across the reported comparisons, pitch errors are generally larger than yaw errors. Our method achieves competitive generalization, with particularly strong results on $D_M$:
our MobileNet model outperforms FSCI with substantially fewer parameters, while our ViT-B model outperforms UniGaze-B using the same backbone architecture. Both comparisons also show improved pitch accuracy on $D_M$.
Nevertheless, these conventional benchmarks provide limited coverage of the challenging illumination and wearable conditions targeted by RealGaze, motivating the complementary robustness evaluations presented in this work.

\subsection{More Ablation Studies}

\paragraph{Variance Loss}

In discretized classification tasks, 
variance loss term is widely used
to encourage a concentrated probability distribution \cite{mean-var}. We evaluate the impact of this term using the recommended weights and a softmax temperature of $\tau=1$.
As shown in \cref{tab:appendix_ablation}(a), the generalization performance between the variance loss and our $\tau=0.5$ technique is similar. However, we advocate for our approach as it is systemically simpler, 
without the need for additional hyperparameter tuning or the computational overhead of auxiliary variance calculations.

\paragraph{Weight Initialization for ViTs}

UniGaze \cite{unigaze} provides 
ViT checkpoints pre-trained on massive unlabeled face datasets via self-supervised learning. We investigate whether these domain-specific weights offer an advantage over those of MAE, which is domain-agnostic.
Results in \cref{tab:appendix_ablation}(b) show that generalization performance remains almost intact regardless of the initialization source. We hypothesize that our SupCon objectives effectively function as a specialized form of self-supervised representation learning. By forcing the model to align features across views and tasks, the ViT backbones efficiently learn gaze-relevant geometric priors, rendering specialized face-only pre-training redundant.

\paragraph{Filtering Training Data}

In \cref{s:setup}, we restrict each training dataset to a shared head pose interval $I_H$ and gaze interval $I$.
To examine the effect of relaxing these restrictions, we train an additional set of models under a near-unfiltered setting, denoted as ``$\simeq0$ filtering''.
Specifically, we remove the head pose constraint and expand the gaze interval to $[-60^\circ, 36^\circ] \times [-60^\circ, 60^\circ]$, with bin sizes $s_\phi=s_\psi=6^\circ$.
We retain this broad gaze interval to support resampling and discretized classification; it excludes less than 2\% of the training data.
As shown in \cref{tab:appendix_ablation}(c), the near-unfiltered models exhibit modest increases in overall error on RealGaze relative to their filtered counterparts; yet our ViT model maintains an advantage over UniGaze-B-Joint in overall RealGaze accuracy.
These results indicate that the proposed method remains effective when trained on a broader range of head poses and gaze directions.

\section{Limitations and Future Work}

While this work significantly advances the generalizability of AGE, it also highlights the persistent challenges inherent in real-world deployment. Our experiments specifically address critical bottlenecks such as facial occlusion and adverse indoor lighting; however, several physiological and environmental factors remain outside the current scope, such as periorbital wrinkles, heavy eye makeup, and extreme lighting conditions like solar over-exposure.
The persistence of these challenges motivates a deeper investigation into the manifold-level representation of ocular features. Furthermore, we hope our findings inspire the development of next-generation synthetic data engines, capable of simulating these edge cases, to bridge the remaining gap between laboratory performance and real-world reliability.

\end{document}